\documentclass[sigconf,screen]{acmart}

\copyrightyear{2022}
\acmYear{2022}
\setcopyright{acmcopyright}
\acmConference[MM '22] {Proceedings of the 30th ACM International Conference on Multimedia }{October 10--14, 2022}{Lisboa, Portugal.}
\acmBooktitle{Proceedings of the 30th ACM International Conference on Multimedia (MM '22), October 10--14, 2022, Lisboa, Portugal}
\acmPrice{15.00}
\acmISBN{978-1-4503-9203-7/22/10}
\acmDOI{10.1145/3503161.3547759}

\usepackage{graphicx}

\usepackage{amsmath}
\usepackage{amssymb}
\usepackage{booktabs}
\usepackage{algorithm}
\usepackage{algpseudocode}
\usepackage{balance}
\algdef{SE}[DOWHILE]{Do}{doWhile}{\algorithmicdo}[1]{\algorithmicwhile\ #1}%

\AtBeginDocument{%
  }

\begin{document}

\title{Im2Oil: Stroke-Based Oil Painting Rendering with Linearly Controllable Fineness Via Adaptive Sampling}


\author{Zhengyan Tong}
\email{418004@sjtu.edu.cn}
\orcid{0000-0003-0376-0972}
\affiliation{%
  \institution{Shanghai Jiao Tong University}
  \country{}
}

\author{Xiaohang Wang}
\email{xygz2014010003@sjtu.edu.cn}
\orcid{0000-0001-5924-7719}
\affiliation{%
  \institution{Shanghai Jiao Tong University}
  \country{}
}

\author{Shengchao Yuan}
\email{sc_yuan@sjtu.edu.cn}
\orcid{0000-0002-8914-489X}
\affiliation{%
  \institution{Shanghai Jiao Tong University}
  \country{}
}

\author{Xuanhong Chen}
\email{chen19910528@sjtu.edu.cn}
\orcid{0000-0002-1072-6804}
\affiliation{%
  \institution{Shanghai Jiao Tong University}
  \country{}
}

\author{Junjie Wang}
\email{dreamboy.gns@sjtu.edu.cn}
\orcid{0000-0002-6368-3879}
\affiliation{%
  \institution{Shanghai Jiao Tong University}
  \country{}
}

\author{Xiangzhong Fang}
\email{xzfang@sjtu.edu.cn}
\orcid{0000-0002-7132-1210}
\affiliation{%
  \institution{Shanghai Jiao Tong University}
  \country{}
}

\renewcommand{\shortauthors}{Zhengyan Tong et al.}

\begin{abstract}
This paper proposes a novel stroke-based rendering (SBR) method that translates images into vivid oil paintings. 
Previous SBR techniques usually formulate the oil painting problem as pixel-wise approximation. Different from this technique route, we treat oil painting creation as an adaptive sampling problem. 
Firstly, we compute a probability density map based on the texture complexity of the input image. Then we use the Voronoi algorithm to sample a set of pixels as the stroke anchors. Next, we search and generate an individual oil stroke at each anchor. Finally, we place all the strokes on the canvas to obtain the oil painting. 
By adjusting the hyper-parameter maximum sampling probability, we can control the oil painting fineness in a linear manner. Comparison with existing state-of-the-art oil painting techniques shows that our results have higher fidelity and more realistic textures. A user opinion test demonstrates that people behave more preference toward our oil paintings than the results of other methods.
More interesting results and the code are in \url{https://github.com/TZYSJTU/Im2Oil}.



\end{abstract}



\begin{teaserfigure}
  \includegraphics[width=\textwidth]{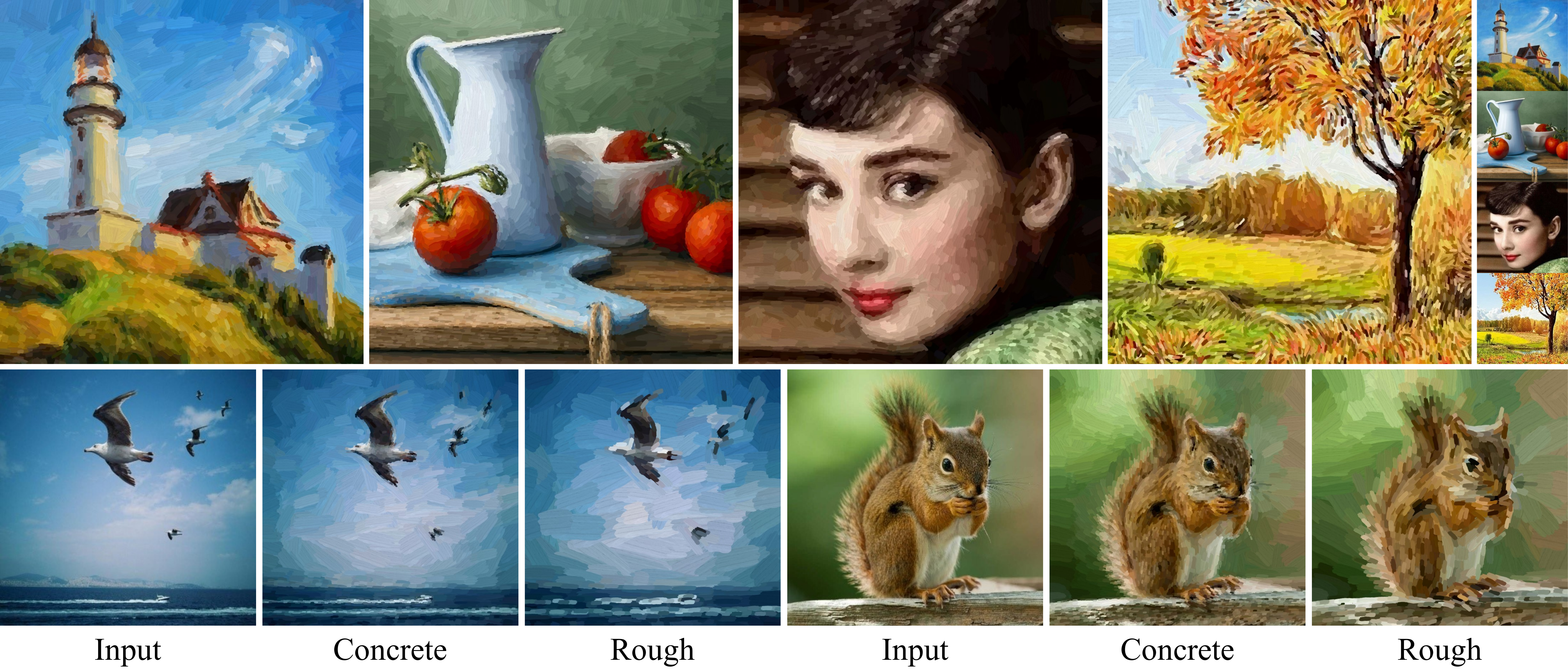}
  \caption{Oil paintings produced by our method. The first row shows four results with the inputs on the right. The second row shows two groups of oil paintings, each including a concrete result and a rough result. We recommend zooming in to see.}
  \Description{Teaser}
  \label{fig:teaser}
\end{teaserfigure}

\maketitle

\section{Introduction}
\label{sec:intro}
Stroke Based Rendering (SBR) \cite{hertzmann2003survey, hertzmann2009stroke} is both a classic topic and a challenging problem in the field of digital multimedia art. Different from the common image-to-image style transfers \cite{Gatys_2016_CVPR, Huang_2017_ICCV, zhu2017unpaired} that are usually based on pixel-wise mapping, SBR techniques create a painting by drawing discrete strokes on the canvas like real artists, which behaves highly intelligent and brings users impressive visual experience. As an imitation of the human creation process, SBR techniques usually follow some heuristic artistic principles, such as from coarse to fine \cite{hertzmann1998painterly, 9010329, liu2021paint} and edge enhancement \cite{10.1145/987657.987676, 10.1145/258734.258893, Tong_Chen_Ni_Wang_2021}. To design an SBR algorithm (this paper focuses on the task of oil painting), researchers need to address three fundamental problems.

The first problem is how to search for strokes. Existing SBR techniques usually formulate oil painting creation as pixel-wise approximation, i.e., minimizing the pixel difference between the oil painting and the input image. The most common methods are greedy strategies, such as \cite{hertzmann1998painterly} drew new strokes at the largest error point. Some other methods \cite{hertzmann2001paint, 10.1145/97880.97902} randomly perturbed the strokes to reduce the sum-of-squares difference. Recent research \cite{9010329} uses deep reinforcement learning to train an oil painting agent by optimizing the $L_2$ distance, while \cite{Zou_2021_CVPR, liu2021paint} train neural networks to predict strokes by optimizing the $L_1$ loss. In this paper, we rethink the oil painting problem and come up with a highly interpretable assumption: areas with complex textures should be painted with small and dense strokes to maintain high fidelity; while areas with smooth textures could be painted with large and sparse strokes since it will not cause obvious visual degradation. Based on this idea, we formulate oil painting creation as an adaptive sampling problem. Firstly, we translate the input image into a probability density map whose probability values are determined by its local texture complexity. Pixels with complex textures have relatively large probabilities, while pixels with smooth textures have relatively small probabilities. Then we use the Voronoi algorithm to sample a set of pixels from the probability density map as the stroke anchors, whose number is proportional to the input image's global texture complexity. At each anchor, we generate an individual stroke based on its nearby texture features. 


The second problem is how to render strokes. We divide existing SBR methods into model rendering and template rendering. Model rendering means the stroke is rendered by some handcrafted math models such as the Bézier curve in \cite{9010329} and the B-spline in \cite{hertzmann2009stroke}. Template rendering means the stroke is derived from a template image, such as the brush dictionary in \cite{zeng2009image} and the primitive brush in \cite{liu2021paint}.
In this paper, we adopt the method of template rendering to produce photo-realistic strokes: we use a gray image of a real oil brush as the stroke template and transform it into various strokes according to some attribute parameters.  


The third problem is how to control the oil painting's fineness.
People usually enjoy fine-grained results, while some others may prefer abstract results. Therefore, SBR techniques usually have a fineness control mechanism: \cite{hertzmann1998painterly} uses a threshold to limit the maximum pixel error of the painting; \cite{10.1145/987657.987676} divides the image into layers of different frequencies; \cite{9010329, Zou_2021_CVPR, liu2021paint} specify the stroke number in each block manually. However, these mechanisms are not user-friendly because they cannot control the fineness in a linear manner. For example, it is hard to predict how much the fineness will improve when reducing the error threshold by 20\% or using 20\% more strokes. Obviously, a linear fineness control mechanism will make it easier for users to tune hyper-parameters and get the desired fineness. In this paper, the fineness of our oil painting is controlled by a hyper-parameter, maximum sampling probability ($p_{max}$), which not only determines the stroke number but also limits the stroke's maximum and minimum size. By adjusting this
hyper-parameter, we can produce oil paintings with linearly changed
fineness from concrete to rough. In Figure \ref{fig:teaser}, the first row shows four concrete results using $p_{max}=\frac{1}{4}$, whose inputs are the small images on the right. The second row shows two groups of oil paintings each including a concrete ($p_{max}=\frac{1}{4}$) result and a rough ($p_{max}=\frac{1}{16}$) result. More discussion about fineness control is in Section \ref{sec:4.2}.

Having addressed all these three problems, we presented this paper and titled it ``Im2Oil'' for short, which means translating an image into oil paintings. To prove the application value of our method, we compare it with three state-of-the-art SBR oil painting techniques \cite{9010329,liu2021paint, Zou_2021_CVPR}. The qualitative comparison shows that our method could generate oil paintings with more realistic textures, higher fidelity, and fewer artifacts.  
For quantitative comparison, we designed a Mean Opinion Score (MOS) test and invited 115 volunteers to rate the oil paintings produced by the above methods.
The MOS test shows there exist obvious gaps among these methods, and the results of our method got the highest user rating.

Our contribution can be summarized as follows: 
\begin{itemize}
\item We rethink the oil painting problem from the perspective of adaptive stroke anchor sampling rather than the commonly followed pixel-wise approximation technique route. 

\item We can linearly control the fineness of our oil painting by varying the hyper-parameter maximum sampling probability ($p_{max}$) to produce arbitrarily concrete or rough results.

\item Both the qualitative comparison and the user opinion test demonstrate that our method is superior to state-of-the-art SBR oil painting techniques.
\end{itemize}

\begin{figure*}[t]
  \centering
   \includegraphics[width=\linewidth]{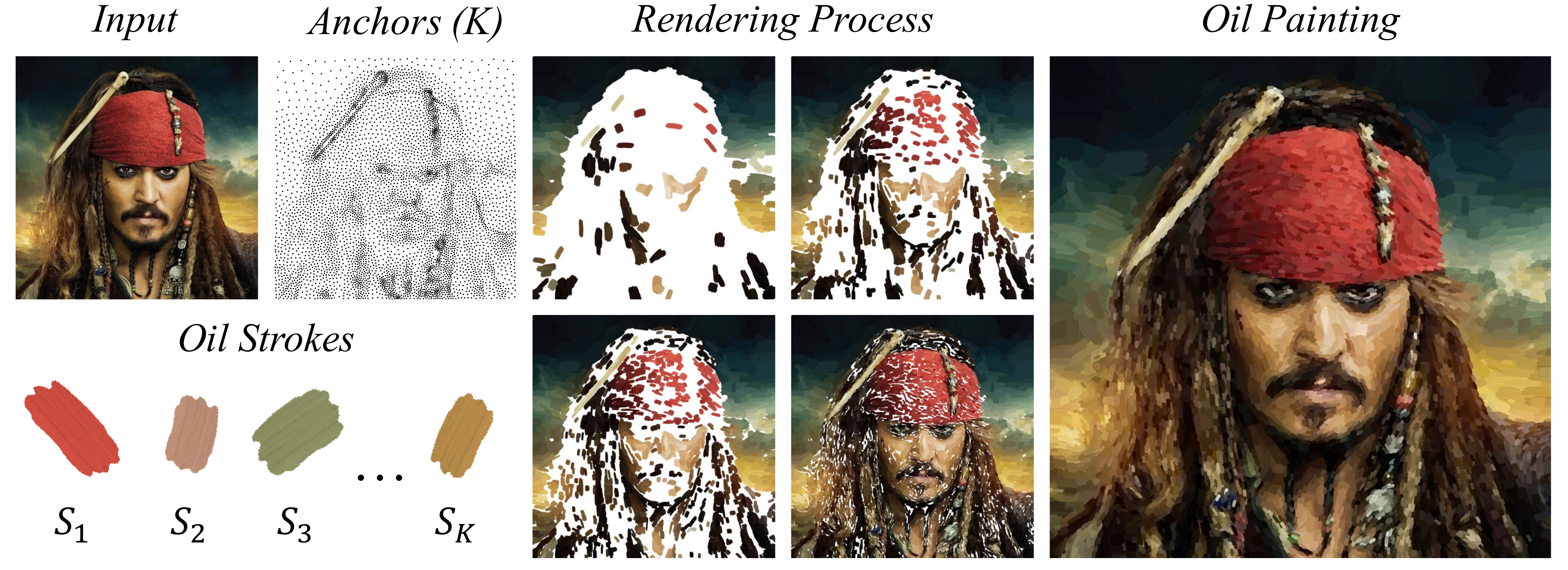}

   \caption{Overall Framework. Firstly, we use the input image to sample a total of K pixels as anchors. Next, we search and generate an individual oil stroke at each anchor, indicated as $\{S_1, S_2, S_3,..., S_K\}$. These strokes are placed one by one on the canvas during the rendering process. Finally, all the strokes compose the oil painting result.}
   \label{fig:Framework}
\end{figure*}

\section{Related Work}
\label{sec:Related Work}

\subsection{Stroke-Based Rendering}
Stroke-Based Rendering (SBR) is a classic artificial intelligence topic that teaches machines to paint like artists. SBR tasks mainly include stippling \cite{Deussen2000Float-6423, 10.1145/508530.508537, 10.1145/1572614.1572622, 1183777, 10.1111/j.1467-8659.2008.01259.x, hiller2003beyond, 1210866, ma2018incremental, vanderhaeghe2007dynamic},
oil painting \cite{10.1145/258734.258893, 10.1145/97880.97902, hertzmann1998painterly, kang2006unified, hertzmann2001paint, 10.1145/508530.508545, 10.1145/1809939.1809948, Zou_2021_CVPR, Kotovenko_2021_CVPR, liu2021paint, 9010329, zeng2009image}, tile mosaics \cite{10.1145/383259.383327, 10.1145/566570.566633, silvers1996photomosaics, 10.1007/BFb0053259, elber2003rendering, di2005artificial, kang2006unified, faustino2005simple}, pen-and-ink drawing \cite{10.1145/192161.192185, 10.1145/192161.192184, 10.1145/258734.258890, winkenbach1996rendering, salisbury1996scale, kang2006unified}, pencil sketching \cite{Tong_Chen_Ni_Wang_2021, ha2018neural, 10.1145/2024156.2024167}, watercolor painting \cite{curtis1997computer, 1309281, kang2006unified, xie2013artist, xie2015stroke}, and vector-field visualization \cite{turk1996image, jobard1997creating, 4015453, 10.1145/984952.984987, healey2001formalizing, laidlaw2001loose}. Technically, SBR algorithms include greedy strategies \cite{10.1145/258734.258893, hertzmann1998painterly, 10.1145/508530.508545, jobard1997creating}, optimization or relaxation \cite{10.1145/508530.508537, Deussen2000Float-6423, 10.1111/j.1467-8659.2008.01259.x, 10.1145/383259.383327}, and trial-and-error methods \cite{10.1145/97880.97902, turk1996image, hertzmann2001paint}. In recent years, researchers have proposed many modern SBR techniques including RL watercolor painting \cite{xie2013artist, xie2015stroke}, RNN-based image doodling \cite{ha2018neural}, CNN-based character writing \cite{zheng2018strokenet}, DRL oil painting \cite{9010329}, Paint Transformer \cite{liu2021paint}, and SBR style transfer \cite{Zou_2021_CVPR, Kotovenko_2021_CVPR}.

\subsection{Voronoi Related SBR Techniques}
In this paper, we use the Voronoi algorithm to sample pixels as the stroke anchors. 
The Voronoi diagram \cite{10.1145/116873.116880} was first applied to the SBR task stippling in \cite{Deussen2000Float-6423} by Deussen \emph{et al.} who use evenly spaced points with different radii to approximate gray images. Inspired by this, Adrian Secord proposed weighted Voronoi stippling \cite{10.1145/508530.508537} by varying the point distribution rather than the point radius. For generalization, Hiller \emph{et al.} \cite{hiller2003beyond} use various small objects with orientation as basic elements of Voronoi stippling.
Kim \emph{et al.} \cite{10.1111/j.1467-8659.2008.01259.x} combined the Voronoi diagram and the Jump Flooding Algorithm to create feature-guided stippling. Ma \emph{et al.} \cite{ma2018incremental} use incremental Voronoi sets for instant stippling. 
Lindemeier \emph{et al.} \cite{lindemeier2015hardware} place new strokes between existing ones by maximizing the minimal distance in the Voronoi diagram. Besides, the Voronoi diagram could also be applied to simulating tile mosaics \cite{elber2003rendering, 10.1145/383259.383327, faustino2005simple}.

\section{Method}
The overall framework of our method is shown in Figure \ref{fig:Framework}. Firstly, we use the \emph{Input} image to sample a total of \emph{K} pixels as \emph{Anchors}. Next, we search and generate an individual \emph{Oil Stroke} at each anchor, indicated as $\{S_1, S_2, S_3,..., S_K\}$. The obtained strokes are placed one by one on the canvas during the \emph{Rendering Process}. Eventually, all the strokes compose the final result \emph{Oil Painting}. In the following, we will introduce our method in detail.

\subsection{Adaptive Anchor Sampling}
\label{sec:3.1}
As introduced in Section \ref{sec:intro}, we vary the stroke size and density in areas of different texture complexity.
Since each stroke corresponds to an individual anchor, controlling the distribution of the strokes is controlling the distribution of their anchors. This section shows how to obtain the stroke anchors by adaptive sampling. Firstly, we use the input image $I$ to calculate a probability density map $I_p$. Pixels in $I_p$ with larger values have a larger probability of being sampled as an anchor. Generally, areas with complex textures (high frequency) are often accompanied by significant gradient changes, while areas with smooth textures (low frequency) usually have small gradients. However, the gradient map is not suitable to directly use as the probability density map because large gradients may only exist on the narrow edges. To paint fine-grained contours, it is not enough to enhance only the edges, but also the neighboring areas of the edges. In other words, we should improve the sampling probability of the pixels around the edges. This can be solved by applying a smoothing operation on the gradient map so that the large gradients will diffuse to their neighboring areas. Specifically, we use the Sobel filter to extract gradients and use the Mean filter to smooth the gradient map. Both filters use a $5\times 5$ window.
Then the smoothed gradient map is normalized to an interval $[p_{min}, p_{max}]$ as the probability density map $I_p$, where $0<p_{min}<p_{max}\leqslant 1$. Here $p_{max}$ is the maximum sampling probability, and $p_{min}$ is the minimum probability. $p_{min}>0$ ensures that the pixels in the most smooth area could be sampled as anchors, so that some strokes will be drawn to this area. $p_{max}\leqslant 1$ means there can be at most one anchor at each pixel. In practice, we use $p_{min}=\frac{1}{100}p_{max}$, thus $p_{max}$ is the only hyper-parameter that need to be specified by users. 
Now add up all the pixel values (sampling probability) in $I_p$, we use the sum $K$ as the number of stroke anchors. Therefore, our method could automatically determine the stoke number based on the global texture complexity of the input image and the hyper-parameter maximum sampling probability $p_{max}$.


\begin{algorithm}[ht] 
    \caption{Rejection Sampling} 
    \label{alg:sampling}
    \renewcommand{\algorithmicrequire}{\textbf{Given:}}
	\renewcommand{\algorithmicensure}{\textbf{Output:}}
    \begin{algorithmic}[1] 
        \Require 
            probability density map $I_{p}$, anchor number $K$;
        \Ensure 
            $K$ anchors;
        \State 
            Randomly generate a probability density map $I_{u}$ which has the same shape as $I_{p}$, and $I_{u}$ $\sim$ Uniform (0,1);
        \State 
            Randomly select a pixel $(x,y)$. If $ I_{u}(x,y)\leqslant I_{p}(x,y)$, sample this pixel; else, reject this pixel.
        \State 
            Repeat Step 2 until sampling $K$ different pixels. 
    \end{algorithmic} 
\end{algorithm}

\begin{figure}[t]
  \centering
   \includegraphics[width=\linewidth]{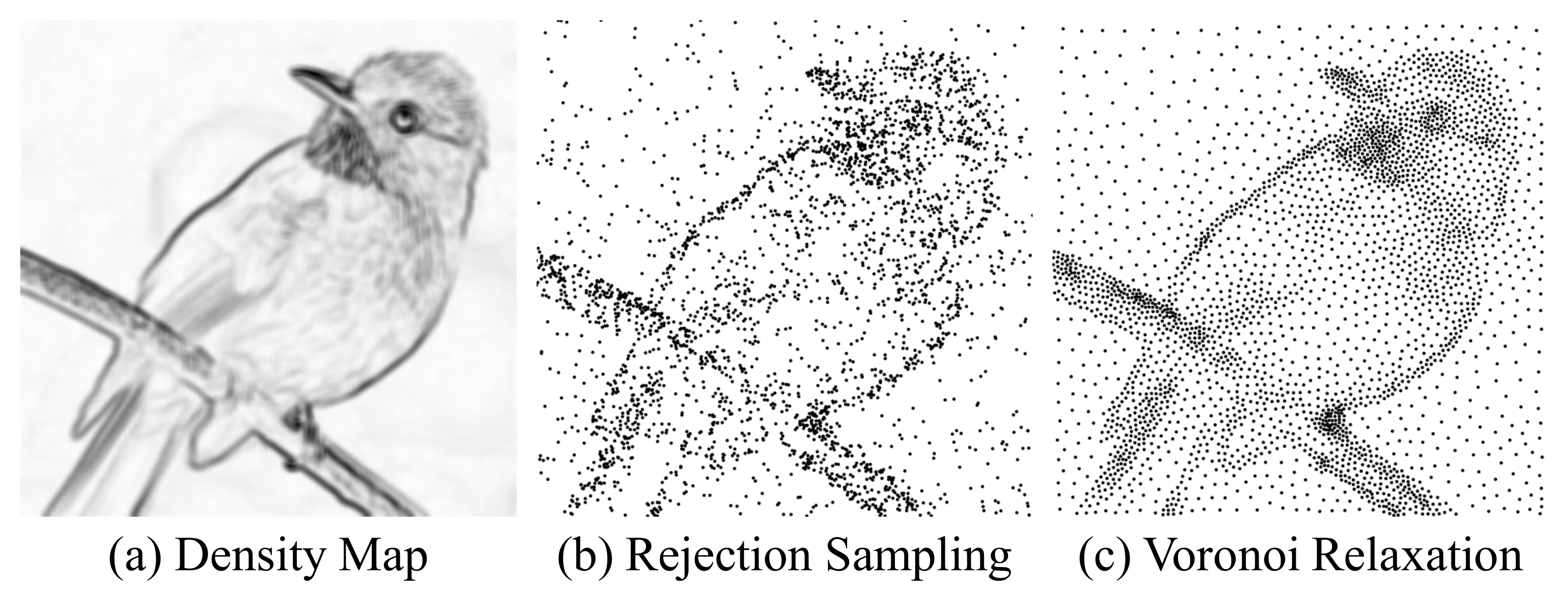}

   \caption{(a) is the probability density map. (b) is the rejection sampling result. We use \textbf{$p_{max}=\frac{1}{4}$} and totally sample \textbf{$K=3068$} anchors. (c) is the Voronoi relaxation result.}
   \label{fig:Voronoi}
\end{figure}

\begin{figure*}[ht]
  \centering
  \includegraphics[width=\linewidth]{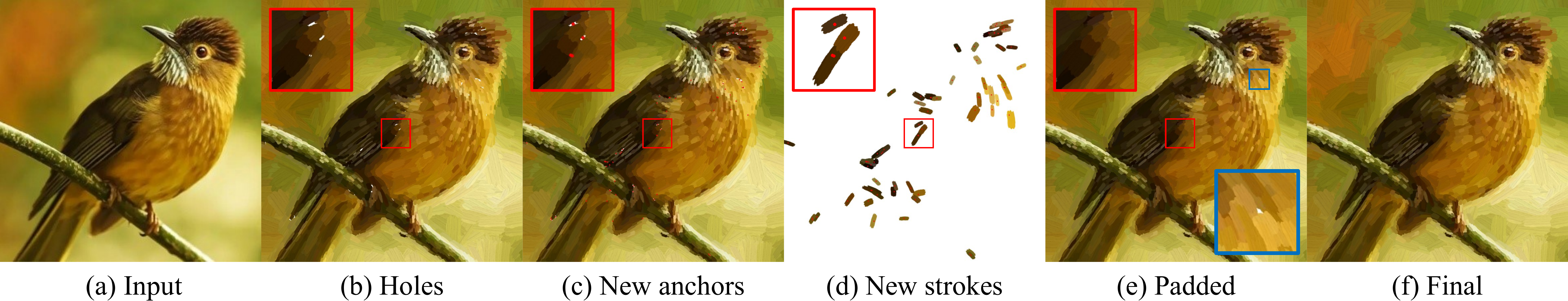}
  \caption{Illustration of the hole padding operation, please zoom in to see. (a) is the input. (b) is the result before padding. In (c), the small red dots are the new anchors. (d) shows the new strokes. (e) is the result of using (d) to pad (b), the zoomed in area with a blue border shows there still exists an uncolored hole. (f) is the final result without any uncolored pixels.}

  \label{fig:Padding}
\end{figure*}

Next, we use Rejection Sampling \cite{mackay1998introduction} to sample $K$ pixels from the probability density map $I_{p}$ as the stroke anchors, as shown in Algorithm \ref{alg:sampling}: we use a uniform distribution $I_{u}$ $\sim$ Uniform (0,1) to sample the probability density map $I_{p}$; every time we randomly select a pixel $(x,y)$, if $I_{u}(x,y)\leqslant I_{p}(x,y)$, then this pixel is sampled (else, rejected); repeat this operation until $K$ different pixels are sampled. By this means, pixels in $I_{p}$ with larger probabilities are more likely to be sampled. In Figure \ref{fig:Voronoi}, (a) is the visualization of a probability density map (the input is in Figure \ref{fig:Padding}). (b) is the rejection sampling result of (a). We use $p_{max}=\frac{1}{4}$, and $K=3068$. The sampled anchors look messy because rejection sampling is just a rough approximation to the probability density map. 
Ideally, the distribution of the anchors should be as close as possible to the probability density map. This can be solved by applying the Voronoi algorithm (Lloyd's method \cite{lloyd1982least}) to optimize (relax) the position of the anchors. Lloyd's method is also known as the famous K-means clustering \cite{bishop1995neural}. As shown in Algorithm \ref{alg:voronoi}: the $K$ anchors obtained by rejection sampling are used as the initial $K$ centroids for pixel clustering; the probability density map $I_{p}$ is used as the weight map of the pixels; each pixel in $I_p$ is divided into the cluster of its nearest (coordinate distance) centroid; the new centroids of these clusters are calculated using the weight map $I_{p}$; repeat the above clustering process for $N$ iterations to optimize the centroid coordinates, which are the final positions of the anchors. In this way, the distribution of the anchors will gradually tend to the probability density map. 
As shown in Figure \ref{fig:Voronoi}, (c) is the Voronoi relaxation result of (b) using Algorithm \ref{alg:voronoi} for $N$=15 iterations. The distribution of the anchors in (c) looks much closer to (a) than in (b). In regions with smooth textures (such as the background), the anchors distribute sparsely and evenly, which mainly depends on the minimum sampling probability $p_{min}$. In regions with significant texture changes (such as the contours), the anchors gather densely while maintaining a certain distance, which mainly depends on the maximum sampling probability $p_{max}$. Actually, for any pixel, assume its sampling probability is $p$, there will be one anchor in its neighboring $\frac{1}{p}$ pixels (approximately). While for any anchor, assume its sampling probability is $p$, there will be no other anchors in its neighboring $\frac{1}{p}$ pixels (approximately). In practice, we fix $N=15$, because after 15 iterations, the anchor distribution will have been close enough to the probability density map.

\begin{algorithm}[ht] 
    \caption{Voronoi Algorithm} 
    \label{alg:voronoi}
    \renewcommand{\algorithmicrequire}{\textbf{Given:}}
	\renewcommand{\algorithmicensure}{\textbf{Output:}}
    \begin{algorithmic}[1] 
        \Require 
            initial $K$ centroids, weight map $I_{p}$, iterations $N$;
        \Ensure 
            optimized $K$ centroids;
        \State 
             For each pixel in $I_{p}$, calculate its coordinate distance to each centroid;
        \State 
             Divide each pixel into the cluster of its nearest centroid; 
        \State 
            Use the weight map $I_{p}$ to calculate the new centroid coordinates of these $K$ clusters; 
        \State 
            Repeat step 1,2,3 for $N$ iterations.        
    \end{algorithmic} 
\end{algorithm}

\begin{figure}[t]
  \centering
   \includegraphics[width=\linewidth]{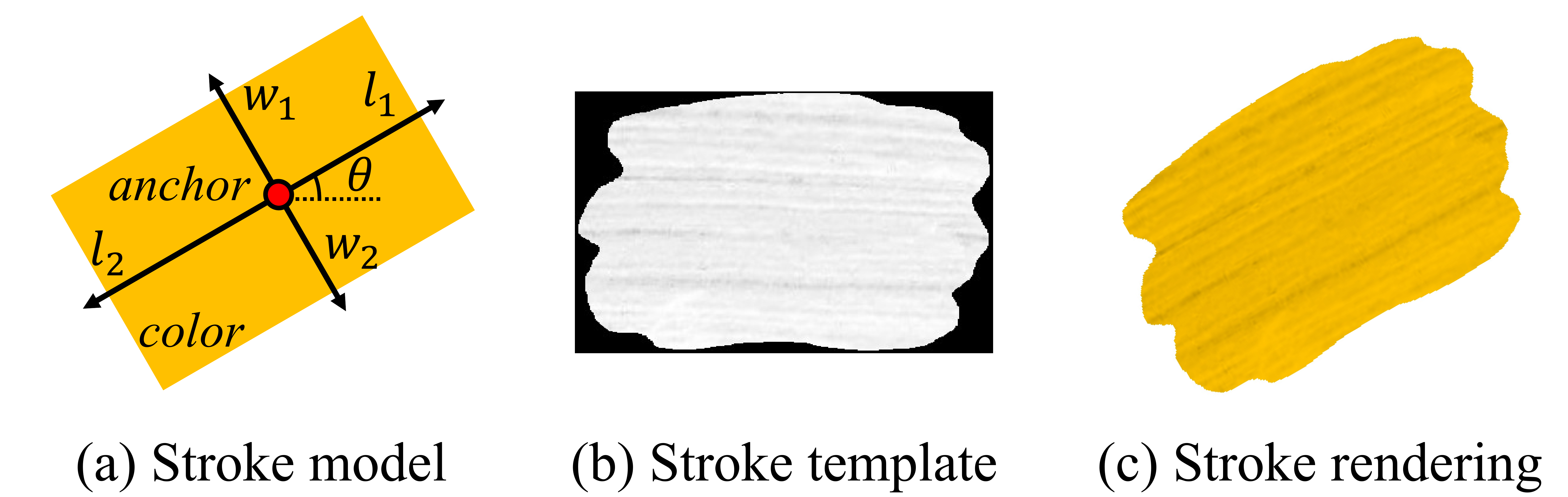}

   \caption{Stroke illustration. (a) is the stroke parameter model. (b) is the stroke template image. (c) is the stroke rendering result using the parameters in (a) and the template in (b).}
   \label{fig:Stroke}
\end{figure}

\begin{algorithm}[h] 
    \caption{Length Searching}    \label{alg:straight}
    \renewcommand{\algorithmicrequire}{\textbf{Input:}}
 \renewcommand{\algorithmicensure}{\textbf{Output:}}
    \begin{algorithmic}[1] 
        \Require 
            anchor coordinate $(x_0,y_0)$, searching direction $\alpha$, input image $(H,S,V)$, H channel threshold $t_H$, V channel threshold $t_V$, step length $\delta$;
        \Ensure 
            length $L$;
        \State $(x,y)\gets (x_0,y_0)$, $L\gets 0$;
        \Do
            \State $L\gets L+\delta$;
            \State $(x,y)\gets (x,y)+(\delta\cdot\cos\alpha, \delta\cdot\sin\alpha)$;
        \doWhile 
            $|H(x,y)-H(x_0,y_0)|< t_H$ and $|V(x,y)-V(x_0,y_0)|< t_V$ 
        \State \Return{$L$}
    \end{algorithmic} 
\end{algorithm}

\subsection{Stroke Searching and Generation}
\label{sec:3.2}
This section introduces how to search and generate a stroke at each anchor. Our stroke model is shown in Figure \ref{fig:Stroke} (a) with the shape of rectangle. The red point indicates the anchor. The rotation angle $\theta$ is the direction of the stroke. $l_1$ is the length extending along the direction of $\theta$; $l_2$ is the length extending along the direction of $(\theta+\pi)$; $w_1$ is the width extending along the direction of $(\theta+\frac{\pi}{2})$; $w_2$ is the width extending along the direction of $(\theta-\frac{\pi}{2})$. Besides, we use the anchor pixel's color as the stroke $color$.

Firstly, we introduce how to determine the stroke direction $\theta$. When human artists paint, they usually draw strokes along the object's edges. That is because drawing strokes across the edges will result in a rough contour \cite{10.1145/258734.258893} while drawing strokes along the edges helps to produce accurate shapes \cite{10.1145/987657.987676}. Therefore, the stroke direction $\theta$ should follow the tangent of the edges. Specifically, we use the ETF vector field (Edge Tangent Flow \cite{10.1145/1274871.1274878, 4522547}) to find the tangent of the edges and use the anchor's ETF direction as $\theta$.
Here briefly reviews the calculation of ETF: compute the input image's gradient vector field (modulus and direction at each pixel); rotate the directions by $\frac{\pi}{2}$; tend the direction of the pixels with a small modulus to the direction of its nearby pixels with a large modulus, while the modulus is not changed. In this way, the ETF direction will approximately follow the tangent of its nearby edges. That is why it calls ``Edge Tangent Flow".


Next, we introduce how to determine the stroke length $l_1$, $l_2$ and width $w_1$, $w_2$, which is described in Algorithm \ref{alg:straight}. We use the anchor coordinate $(x_0,y_0)$ as the start point for searching. The searching direction $\alpha=\theta,\theta+\pi,\theta+\frac{\pi}{2},\theta-\frac{\pi}{2}$ for $l_1,l_2,w_1,w_2$, respectively. The input image is converted to the HSV color space since HSV (Hue, Saturation, Value) have more intuitive meanings than RGB. In every step, the point $(x,y)$ moves length $\delta$ along the searching direction $\alpha$ (we fix $\delta$ = 1 pixel).
During searching, two conditions constrain the difference between the point $(x,y)$ and the anchor $(x_0,y_0)$. $|H(x,y)-H(x_0,y_0)|<t_H$ restricts the stroke from being drawn to regions with much different hue. $|V(x,y)-V(x_0,y_0)|< t_V$  restricts the stroke from being drawn to regions with much different brightness. In practice, we fix $t_H=\frac{\pi}{3}$ and $t_V=15$ as the thresholds. 
We do not set a condition for the S channel because the visual difference caused by saturation is not as obvious as hue and brightness. Besides, we use $L_{min}$ and $L_{max}$ as the minimum and maximum length to clip the searching length $L$. $L_{min}$ and $L_{max}$ are automatically determined by the anchor's sampling probability. As described at the end of Section \ref{sec:3.1}, assume the anchor's sampling probability is $p$, there will be no other anchors in its neighboring $\frac{1}{p}$ pixels (approximately), whose area equals to a $p^{-\frac{1}{2}}\times p^{-\frac{1}{2}}$ square window. In view of this, it is reasonable to make $L_{max}$ proportional to $p^{-\frac{1}{2}}$. In practice, we use $L_{max}$ = $p^{-\frac{1}{2}}$ for $w_1$ and $w_2$, and we use $L_{max}$ = $3\times p^{-\frac{1}{2}}$ for $l_1$ and $l_2$. That's because the stroke length are generally larger than width. As for $L_{min}$, we use $L_{min}$ = ${p_{max}}^{-\frac{1}{2}}$ for $w_1$ and $w_2$ and $L_{min}$ = $3\times {p_{max}}^{-\frac{1}{2}}$ for $l_1$ and $l_2$. $L_{min}$ limits the fidelity (detail resolution) that the oil painting can achieve, which essentially depends on the maximum sampling probability $p_{max}$.


As introduced in Section \ref{sec:intro}, we use the method of template to render strokes. Figure \ref{fig:Stroke} (b) shows our oil stroke template, which is a gray image composed of a background mask and a foreground texture. Only the foreground is the actually used part. The mask with such an irregular shape looks much more natural than a rigid rectangle. The texture follows the long side of the stroke to reflect the drawing direction. Now given the stroke parameters $\theta$, $l_1$, $l_2$, $w_1$, $w_2$, and HSV color $(h,s,v)$, the stroke template is resized to length $l_1+l_2$, width $w_1+w_2$, and is rotated by angle $\theta$. Then the template is colored by the following method. Take constant $h$ and $s$ as the stroke's H channel and S channel. For the V channel, the value varies with the texture of the template. Use $T$ to denote the template image and use
$g_m$ to denote its grayscale average. The template $T$ is multiplied by $\frac{v}{g_m}$ as the V channel of the stroke. In this way, the mean energy of the V channel will not change because $Mean(\frac{v}{g_m}T)=\frac{v}{g_m}Mean(T)=v$. 
As shown in Figure \ref{fig:Stroke}, (c) is the stroke rendering result using the parameters in (a) and the template in (b). In addition to the stroke template in Figure \ref{fig:Stroke} (b), we also offer a set of alternative templates with different textures, which is discussed in the \href{https://github.com/TZYSJTU/Im2Oil}{supplementary material}.

\begin{figure*}[ht]
  \centering
  \includegraphics[width=\linewidth]{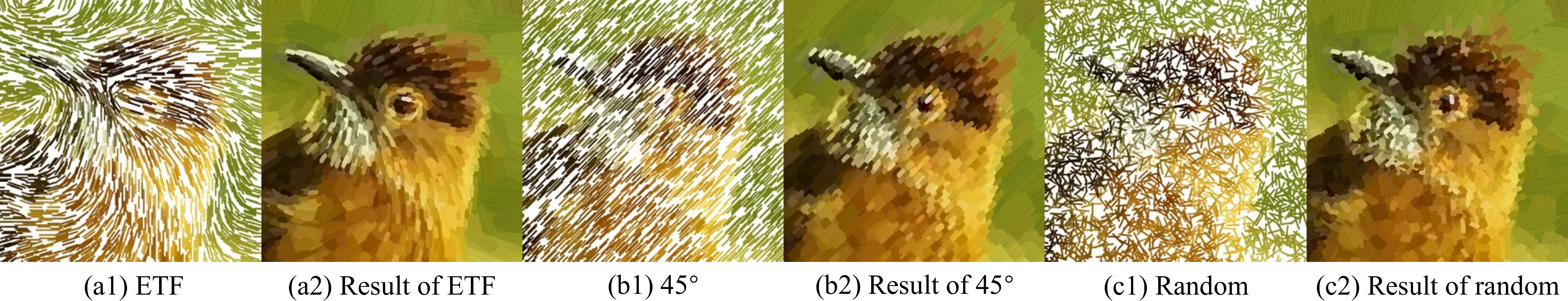}
  \caption{Comparison of different stroke direction. (a1) is the illustration of the ETF vector field; (b1) is the illustration of a constant vector field fixed at $45^{\circ}$; (c1) is the illustration of a random vector field; (a2)(b2)(c2) are the corresponding oil painting results whose strokes are guided by the direction of (a1)(b1)(c1), respectively.}

  \label{fig:ETF}
\end{figure*}

\begin{figure*}[ht]
  \centering
  \includegraphics[width=\linewidth]{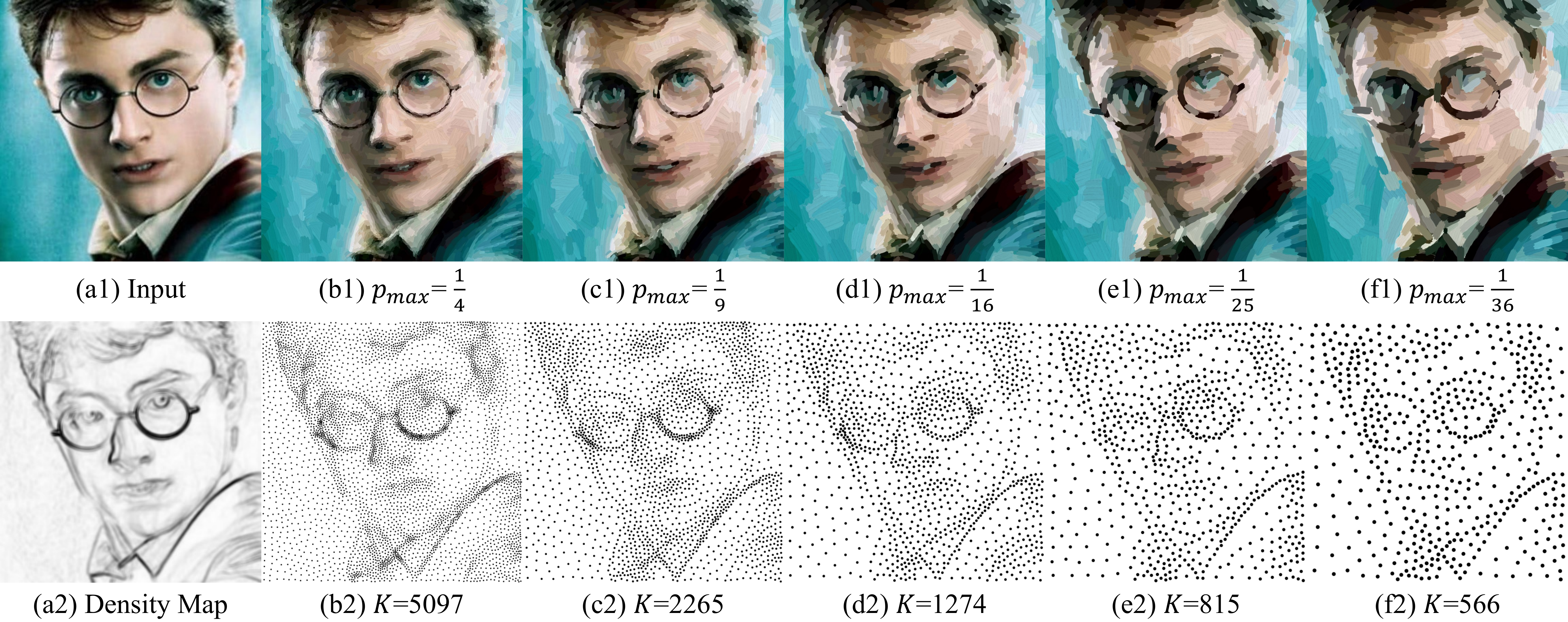}
  \caption{Illustration of oil painting fineness control. (a1) is the input; (a2) is the probability density map; (b1)(c1)(d1)(e1)(f1) are the oil painting result using $p_{max}=\{\frac{1}{4}, \frac{1}{9}, \frac{1}{16}, \frac{1}{25}, \frac{1}{36}\}$, respectively; (b2)(c2)(d2)(e2)(f2) are the corresponding anchors of (b1)(c1)(d1)(e1)(f1), which samples $K=\{5097,2265,1274,815,566\}$ anchors, respectively.}

  \label{fig:fineness}
\end{figure*}

\subsection{Painting Process}
\label{sec:3.3}
By the methods in Section \ref{sec:3.2}, we could search and generate a stroke at each anchor. Now we place the strokes one by one on the canvas to produce the oil painting. During this process, some strokes will overlap with each other. In this case, we make the newly drawn stroke completely cover the previous ones, which is consistent with the real oil painting. Therefore, different stroke drawing orders will lead to different final results. If the stroke number is $K$, then we have $K!$ different orders, and there must be an optimal solution that minimizes the pixel difference between the painting result and the original image. However, the factorial of $K$ is too large to enumerate each order and find the optimal solution. Fortunately, we can quickly find an approximate solution. Review the anchor sampling method in Section \ref{sec:3.1} and the stroke length clipping mechanism in Section \ref{sec:3.2}. Areas with complex textures are sampled with dense anchors and drawn with small strokes, while areas with smooth textures are sampled with sparse anchors and drawn with large strokes. Obviously, large strokes are more likely to cause errors, while small strokes are usually more accurate. In view of this, we sort the strokes by their area size and paint them from the largest to the smallest (greedy strategy).

As shown in Figure \ref{fig:Padding}, (a) is the input image and (b) is the oil painting result after adding all the strokes to the canvas. Zoom in (b), it can be seen that a few small white holes are not colored. The enlarged area with a red border shows some of the holes. One reason for this phenomenon is that the mask of the stroke template discards some border regions and these regions happen to be not colored by any other strokes during the painting process. Besides, the noise in the image may influence stroke searching and cause some holes. Whatever, the current algorithm cannot guarantee to paint all the pixels. 
We solve this problem by the following padding method. 
Firstly, we find the connected domain of each uncolored hole and calculate the centroid of the connected domain.
As shown in Figure \ref{fig:Padding} (c), the tiny red points are the centroids of the uncolored holes. Next, we use these centroids as new anchors to search and render strokes on a white canvas, obtaining (d). The tiny red points in (d) are the new anchors, whose positions are the same as those in (c). Now use (d) to pad (b), which means the value of those uncolored pixels in (b) are taken from the corresponding pixels in (d), while the value of those already colored pixels in (b) remain unchanged. The padding result is shown in (e). Although most of the holes disappear, the padding operation may not eliminate all the uncolored pixels at one time. The enlarged area with a blue border in (e) shows that one hole still exists. Therefore, the padding operation will usually iterate several times (empirically, three times in most cases) before eliminating all the holes. Finally, we get the oil painting result with no uncolored pixels, as shown in (f).

\section{Ablation Study}

\subsection{Stroke Direction}

As described in Section \ref{sec:3.2}, we use the Edge Tangent Flow vector field \cite{10.1145/1274871.1274878, 4522547} as the stroke direction guidance. The ablation study of ETF is shown in Figure \ref{fig:ETF}: (a1) is the visualization of the ETF direction, where the short lines denote the local ETF direction; (a2) is the oil painting result using ETF as its stroke direction guidance; (b1) is the visualization of a constant vector field with the direction of $45^{\circ}$; (b2) is the oil painting result using $45^{\circ}$ as its stroke direction guidance; (c1) is the visualization of a random vector field; (c2) is the oil painting result using this random vector field as its stroke direction guidance. 
Observing (a2)(b2)(c2), the effectiveness of ETF reflects in two aspects. On the one hand, ETF guides the strokes along the tangent of the edges, which helps to produce sharp edges and clear details. For example, the bird's beak in (a2) has much sharper edges than in (b2)(c2), and the bird's eye in (a2) has more accurate details than in (b2) and (c2). On the other hand, ETF can capture the texture orientation features, which helps to preserve the content's semantic information. For example, the strokes of the bird's feathers in (a2) follow the feathers' actual growth direction, while the strokes in (b2) and (c2) cannot reflect the feathers' orientation feature. Both of these illustrate that using ETF could make the oil paining results more aesthetic.

\begin{figure*}[ht]
  \centering
  \includegraphics[width=\linewidth]{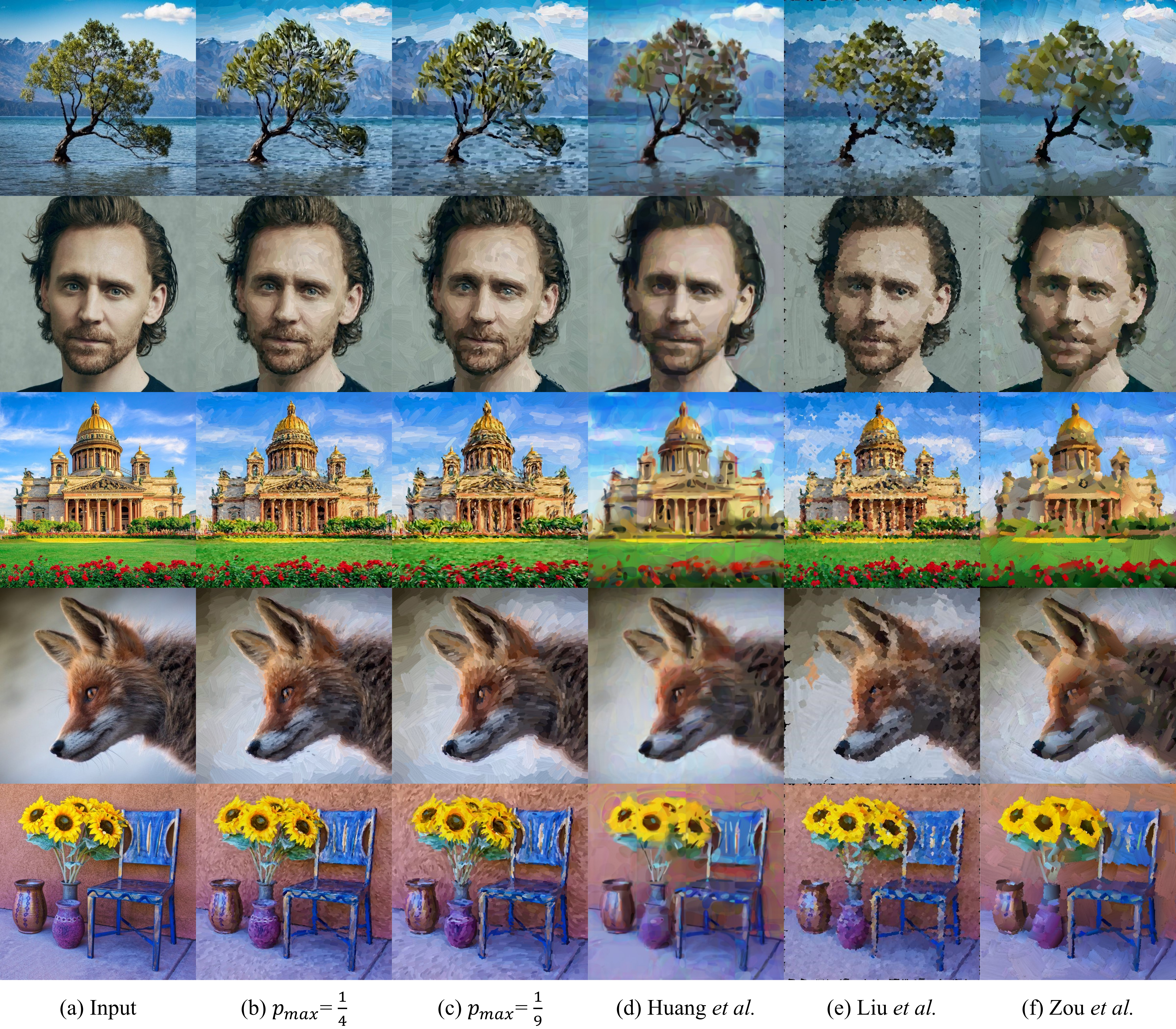}
  \caption{
    Oil painting comparison with three state-of-the-art methods: Huang \emph{et al.} \cite{9010329}, Liu \emph{et al.} \cite{liu2021paint}, and Zou \emph{et al.} \cite{Zou_2021_CVPR}. Column (a) is the input. Column (b) and (c) are our results using $p_{max}=\{\frac{1}{4}, \frac{1}{9}\}$, respectively. Column (d) is the result of Huang \emph{et al.} \cite{9010329}; (e) is the result of Liu \emph{et al.} \cite{liu2021paint}; (f) is the result of Zou \emph{et al.} \cite{Zou_2021_CVPR}. Please zoom in to compare details.}
  \label{fig:compare}
\end{figure*}

\subsection{Fineness Control}
\label{sec:4.2}
We use the hyper-parameter maximum sampling probability $p_{max}$ to control the fineness of our oil painting explicitly and linearly. As shown in Figure \ref{fig:fineness}: (a1) is the input image; (a2) is the probability density map; (b1)(c1)(d1)(e1)(f1) are the oil painting results using $p_{max}=\{\frac{1}{4}, \frac{1}{9}, \frac{1}{16}, \frac{1}{25}, \frac{1}{36}\}$, respectively; (b2)(c2)(d2)(e2)(f2) are the corresponding anchors of (b1)(c1)(d1)(e1)(f1), which samples $K=\{5097,2265,1274,815,566\}$ anchors, respectively. Actually, the anchor number $K$ is proportional to $p_{max}$ due to the normalization operation in Section \ref{sec:3.1}. 
We use value $\{\frac{1}{4}, \frac{1}{9}, \frac{1}{16}, \frac{1}{25}, \frac{1}{36}\}$ for $p_{max}$ because the maximum sampling density corresponds to one anchor in a square window with the side length of $\{2, 3, 4, 5, 6\}$, respectively. Therefore, the fineness of the oil painting changes linearly from (b1) to (f1), which can be seen from the glasses and the hair. More clearly, the fineness is proportional to ${p_{max}}^{-\frac{1}{2}}$.
Using $p_{max}=\frac{1}{4}$ could produce fine-grained results while using $p_{max}=\frac{1}{36}$ leads to a very abstract result. For users who don't know the technical details, we just provide these five fineness levels for them to choose from, which is enough to meet the needs of application scenarios. Besides, it should be noted that our fineness control method is content-free, which means the fineness only depends on the parameter $p_{max}$, regardless of the input image. This is because $p_{max}$ determines the maximum density of strokes, while the image content only affects the total number of strokes.

\section{Comparison Experiment}

\subsection{Qualitative Comparison}
We compare our method with three state-of-the-art SBR oil painting techniques. As shown in Figure \ref{fig:compare}: column (a) is the input image; column (b)(c) are our results using $p_{max}=\{\frac{1}{4},\frac{1}{9}\}$, respectively; (d)(e)(f) are the results of Huang \emph{et al.} \cite{9010329}, Liu \emph{et al.} \cite{liu2021paint}, and Zou \emph{et al.} \cite{Zou_2021_CVPR}, respectively. The results of the compared methods are generated by their official codes with default parameters.

Huang \emph{et al.} \cite{9010329} use the deep reinforcement learning algorithm DDPG (Deep Deterministic Policy Gradient) to train an oil painting agent. The main weakness of this method is that the results are too blurred and lack style since its oil painting agent is trained by minimizing the $L2$ loss. Besides, this method uses Bézier curves as its stroke model and merges the strokes by transparency (alpha channel), both of which are inconsistent with actual oil painting, making its textures look unreal, and the overlapping areas appear ghosting-like artifacts. Compared with Huang \emph{et al.} \cite{9010329}, our oil paintings are distinctive in style and have no artifacts. 
Liu \emph{et al.} \cite{liu2021paint} devise a self-training network to produce oil paintings by predicting the parameters of a stroke set. The main weakness of this method is that the results look too rough, failing to form sharp edges. Most of the strokes are square and fragmentary, making the results like mosaics. In addition, this method somehow cannot fill up the entire canvas. There exist a lot of uncolored small black areas, especially at the borders. Compared with Liu \emph{et al.} \cite{liu2021paint}, our method achieves higher fidelity, and we can guarantee to fill up the canvas by the padding method introduced in Section \ref{sec:3.3}. 
Zou \emph{et al.} \cite{Zou_2021_CVPR} design a rasterization network with a shading network to generate oil paintings. Similar to Huang \emph{et al.}, the main weakness of Zou \emph{et al.}'s method is that the results look too blurred, failing to preserve tiny details. Moreover, the stroke boundaries exist many obvious noise points, which heavily degrade the visual quality. Compared with Zou \emph{et al.} \cite{Zou_2021_CVPR}, we can produce more detailed results, and our texture looks cleaner and clearer.

\subsection{Quantitative Comparison}
\label{sec:Quantitative}
To further prove the application value of our method, we organized a Mean Opinion Score (MOS) test to measure the user preference for the above oil painting methods. Firstly, we collected a gallery dataset with fifty input images, which can be divided into five sets: scenery, portrait, animal, building, and still life. Every set contains ten images. For each image in this dataset, we use the compared methods \cite{9010329}\cite{liu2021paint}\cite{Zou_2021_CVPR} and ours with $p_{max}=\{\frac{1}{4},\frac{1}{9}\}$ to produce five oil painting results.

Next, we invited a total of 115 volunteers from our university to participate in the MOS test. To avoid opinion bias among these participants, they were from eight different majors, each in roughly equal numbers. The volunteers were asked to score the oil paintings based on their visual perception in the following manner: every input image and its five oil paintings were packaged together as a group for display; the five oil paintings were anonymous in every group, and their sequence was shuffled in different groups; participants had to observe each group of oil paintings for at least 10 seconds before giving their opinion scores; the available score values were integers ranging from 1 to 10. Besides, three groups of oil paintings would appear twice in the MOS test. If all these three groups of oil paintings' score rankings differ in the first and the second time, then we have reason to suspect that this participant is unreliable. Opinion scores collected from unreliable participants would be discarded. By this means, we identified 7 unreliable participants among these 115 volunteers.

After all the participants gave their opinion scores, for every reliable participant, his/her scores were normalized
to $mean = 7.5$ and $variance = 1.5$. Finally, we calculated the average score for each method on each image set. The average opinion scores of different methods are shown in Table \ref{tab:MOS}. Set 1-5 correspond to scenery, portrait, animal, building, and still life, respectively. The last column is the average score over the whole dataset. In each column, the best score is in bold, and the second is underlined. As a result, ours using $p_{max}=\frac{1}{4}$ achieves the highest opinion score in each set and the whole dataset; Huang \emph{et al.} \cite{9010329} achieves the second in Set-3; Liu \emph{et al.} \cite{liu2021paint} achieves the second in Set-4; Zou \emph{et al.} \cite{Zou_2021_CVPR} achieves the second in Set-2; ours using $p_{max}=\frac{1}{9}$ achieves the second in Set-1, Set-5, and the whole dataset. In the last column, the mean scores of different methods have apparent gaps. Overall, users behave more preference to our oil paintings than the results of the compared techniques. The dataset and the corresponding oil painting results can be found in our \href{https://github.com/TZYSJTU/Im2Oil}{supplementary material}.

\begin{table}[ht]
  \caption{MOS test. Set 1-5 correspond to scenery, portrait, animal, building, and still life, respectively. In every column, the best score is in bold and the second is underlined.}
  \label{tab:MOS}
  \begin{tabular}{@{}lcccccc@{}}
    \toprule
    Methods & Set-1  &  Set-2  &  Set-3  & Set-4  & Set-5  & Mean \\
    \midrule
    
    Huang \emph{et al.} \cite{9010329}  & 6.91  & 7.42  & \underline{7.60}  & 6.88  & 7.06  & 7.17\\
    
    Liu \emph{et al.} \cite{liu2021paint}   & 7.32  & 6.58  & 6.74  & \underline{7.47} & 6.69  & 6.96\\
    
    Zou \emph{et al.}  \cite{Zou_2021_CVPR}  & 7.12  & \underline{8.07}  & 7.28 & 7.25 & 7.58  & 7.46\\  
    
    Ours, $p_{max}=\frac{1}{9}$  & \underline{7.98}  & 7.61  & 7.44  & 7.37 & \underline{7.84} & \underline{7.65}\\
    Ours, $p_{max}=\frac{1}{4}$  & \textbf{8.46}  & \textbf{8.27}  & \textbf{8.20}   & \textbf{8.03} & \textbf{8.34}  & \textbf{8.26}\\
    \bottomrule
  \end{tabular}
\end{table}

\section{Conclusion and Future Work}
In this paper, we rethink the SBR oil painting problem from the perspective of adaptive stroke anchor sampling rather than the common pixel-wise approximation technique route. The anchor density is determined by the local texture complexity; the anchor number is determined by the global texture complexity; the anchor position is optimized by the Voronoi algorithm. By adjusting the hyper-parameter maximum sampling probability, we can control the oil painting's fineness from concrete to rough in a linear manner. Visual comparison with state-of-the-art SBR oil painting techniques demonstrates that our method achieves higher fidelity and better texture quality. The MOS test shows users behave more preference to our oil paintings than the results of other methods. In the future, we will try to extend our sampling-based method to more SBR styles, such as watercolor painting and pencil sketching. We may also explore the SBR style transfer problem.




\bibliographystyle{ACM-Reference-Format}
\balance
\bibliography{Paper79}


\begin{thebibliography}{60}


\ifx \showCODEN    \undefined \def \showCODEN     #1{\unskip}     \fi
\ifx \showDOI      \undefined \def \showDOI       #1{#1}\fi
\ifx \showISBNx    \undefined \def \showISBNx     #1{\unskip}     \fi
\ifx \showISBNxiii \undefined \def \showISBNxiii  #1{\unskip}     \fi
\ifx \showISSN     \undefined \def \showISSN      #1{\unskip}     \fi
\ifx \showLCCN     \undefined \def \showLCCN      #1{\unskip}     \fi
\ifx \shownote     \undefined \def \shownote      #1{#1}          \fi
\ifx \showarticletitle \undefined \def \showarticletitle #1{#1}   \fi
\ifx \showURL      \undefined \def \showURL       {\relax}        \fi
\providecommand\bibfield[2]{#2}
\providecommand\bibinfo[2]{#2}
\providecommand\natexlab[1]{#1}
\providecommand\showeprint[2][]{arXiv:#2}

\bibitem[Aurenhammer(1991)]%
        {10.1145/116873.116880}
\bibfield{author}{\bibinfo{person}{Franz Aurenhammer}.}
  \bibinfo{year}{1991}\natexlab{}.
\newblock \showarticletitle{Voronoi Diagrams—a Survey of a Fundamental
  Geometric Data Structure}.
\newblock \bibinfo{journal}{\emph{ACM Comput. Surv.}} \bibinfo{volume}{23},
  \bibinfo{number}{3} (\bibinfo{date}{Sept.} \bibinfo{year}{1991}),
  \bibinfo{pages}{345–405}.
\newblock
\showISSN{0360-0300}
\urldef\tempurl%
\url{https://doi.org/10.1145/116873.116880}
\showDOI{\tempurl}


\bibitem[Bishop et~al\mbox{.}(1995)]%
        {bishop1995neural}
\bibfield{author}{\bibinfo{person}{Christopher~M Bishop} {et~al\mbox{.}}}
  \bibinfo{year}{1995}\natexlab{}.
\newblock \bibinfo{booktitle}{\emph{Neural networks for pattern recognition}}.
\newblock \bibinfo{publisher}{Oxford university press}.
\newblock


\bibitem[Curtis et~al\mbox{.}(1997)]%
        {curtis1997computer}
\bibfield{author}{\bibinfo{person}{Cassidy~J Curtis}, \bibinfo{person}{Sean~E
  Anderson}, \bibinfo{person}{Joshua~E Seims}, \bibinfo{person}{Kurt~W
  Fleischer}, {and} \bibinfo{person}{David~H Salesin}.}
  \bibinfo{year}{1997}\natexlab{}.
\newblock \showarticletitle{Computer-generated watercolor}. In
  \bibinfo{booktitle}{\emph{Proceedings of the 24th annual conference on
  Computer graphics and interactive techniques}}. \bibinfo{pages}{421--430}.
\newblock


\bibitem[Deussen et~al\mbox{.}(2000)]%
        {Deussen2000Float-6423}
\bibfield{author}{\bibinfo{person}{Oliver Deussen}, \bibinfo{person}{Stefan
  Hiller}, \bibinfo{person}{Cornelius~van Overveld}, {and}
  \bibinfo{person}{Thomas Strothotte}.} \bibinfo{year}{2000}\natexlab{}.
\newblock \showarticletitle{Floating Points: A Method for Computing Stipple
  Drawings}.
\newblock \bibinfo{journal}{\emph{Computer Graphics Forum}}
  \bibinfo{volume}{19}, \bibinfo{number}{3} (\bibinfo{year}{2000}).
\newblock
\urldef\tempurl%
\url{https://doi.org/10.1111/1467-8659.00396}
\showDOI{\tempurl}


\bibitem[Di~Blasi and Gallo(2005)]%
        {di2005artificial}
\bibfield{author}{\bibinfo{person}{Gianpiero Di~Blasi} {and}
  \bibinfo{person}{Giovanni Gallo}.} \bibinfo{year}{2005}\natexlab{}.
\newblock \showarticletitle{Artificial mosaics}.
\newblock \bibinfo{journal}{\emph{The Visual Computer}} \bibinfo{volume}{21},
  \bibinfo{number}{6} (\bibinfo{year}{2005}), \bibinfo{pages}{373--383}.
\newblock


\bibitem[Elber and Wolberg(2003)]%
        {elber2003rendering}
\bibfield{author}{\bibinfo{person}{Gershon Elber} {and} \bibinfo{person}{George
  Wolberg}.} \bibinfo{year}{2003}\natexlab{}.
\newblock \showarticletitle{Rendering traditional mosaics}.
\newblock \bibinfo{journal}{\emph{The Visual Computer}} \bibinfo{volume}{19},
  \bibinfo{number}{1} (\bibinfo{year}{2003}), \bibinfo{pages}{67--78}.
\newblock


\bibitem[Faustino and de~Figueiredo(2005)]%
        {faustino2005simple}
\bibfield{author}{\bibinfo{person}{Geisa~Martins Faustino} {and}
  \bibinfo{person}{Luiz~Henrique de Figueiredo}.}
  \bibinfo{year}{2005}\natexlab{}.
\newblock \showarticletitle{Simple adaptive mosaic effects}. In
  \bibinfo{booktitle}{\emph{XVIII Brazilian Symposium on Computer Graphics and
  Image Processing (SIBGRAPI'05)}}. IEEE, \bibinfo{pages}{315--322}.
\newblock


\bibitem[Finkelstein and Range(1998)]%
        {10.1007/BFb0053259}
\bibfield{author}{\bibinfo{person}{Adam Finkelstein} {and}
  \bibinfo{person}{Marisa Range}.} \bibinfo{year}{1998}\natexlab{}.
\newblock \showarticletitle{Image mosaics}. In
  \bibinfo{booktitle}{\emph{Electronic Publishing, Artistic Imaging, and
  Digital Typography}}, \bibfield{editor}{\bibinfo{person}{Roger~D. Hersch},
  \bibinfo{person}{Jacques Andr{\'e}}, {and} \bibinfo{person}{Heather Brown}}
  (Eds.). \bibinfo{publisher}{Springer Berlin Heidelberg},
  \bibinfo{address}{Berlin, Heidelberg}, \bibinfo{pages}{11--22}.
\newblock
\showISBNx{978-3-540-69718-3}


\bibitem[Fu et~al\mbox{.}(2011)]%
        {10.1145/2024156.2024167}
\bibfield{author}{\bibinfo{person}{Hongbo Fu}, \bibinfo{person}{Shizhe Zhou},
  \bibinfo{person}{Ligang Liu}, {and} \bibinfo{person}{Niloy~J. Mitra}.}
  \bibinfo{year}{2011}\natexlab{}.
\newblock \showarticletitle{Animated Construction of Line Drawings}. In
  \bibinfo{booktitle}{\emph{Proceedings of the 2011 SIGGRAPH Asia Conference}}
  (Hong Kong, China) \emph{(\bibinfo{series}{SA '11})}.
  \bibinfo{publisher}{Association for Computing Machinery},
  \bibinfo{address}{New York, NY, USA}, Article \bibinfo{articleno}{133},
  \bibinfo{numpages}{10}~pages.
\newblock
\showISBNx{9781450308076}
\urldef\tempurl%
\url{https://doi.org/10.1145/2024156.2024167}
\showDOI{\tempurl}


\bibitem[Gatys et~al\mbox{.}(2016)]%
        {Gatys_2016_CVPR}
\bibfield{author}{\bibinfo{person}{Leon~A. Gatys},
  \bibinfo{person}{Alexander~S. Ecker}, {and} \bibinfo{person}{Matthias
  Bethge}.} \bibinfo{year}{2016}\natexlab{}.
\newblock \showarticletitle{Image Style Transfer Using Convolutional Neural
  Networks}. In \bibinfo{booktitle}{\emph{Proceedings of the IEEE Conference on
  Computer Vision and Pattern Recognition (CVPR)}}.
\newblock


\bibitem[Gooch et~al\mbox{.}(2002)]%
        {10.1145/508530.508545}
\bibfield{author}{\bibinfo{person}{Bruce Gooch}, \bibinfo{person}{Greg Coombe},
  {and} \bibinfo{person}{Peter Shirley}.} \bibinfo{year}{2002}\natexlab{}.
\newblock \showarticletitle{Artistic Vision: Painterly Rendering Using Computer
  Vision Techniques}. In \bibinfo{booktitle}{\emph{Proceedings of the 2nd
  International Symposium on Non-Photorealistic Animation and Rendering}}
  (Annecy, France) \emph{(\bibinfo{series}{NPAR '02})}.
  \bibinfo{publisher}{Association for Computing Machinery},
  \bibinfo{address}{New York, NY, USA}, \bibinfo{pages}{83–ff}.
\newblock
\showISBNx{1581134940}
\urldef\tempurl%
\url{https://doi.org/10.1145/508530.508545}
\showDOI{\tempurl}


\bibitem[Ha and Eck(2018)]%
        {ha2018neural}
\bibfield{author}{\bibinfo{person}{David Ha} {and} \bibinfo{person}{Douglas
  Eck}.} \bibinfo{year}{2018}\natexlab{}.
\newblock \showarticletitle{A Neural Representation of Sketch Drawings}. In
  \bibinfo{booktitle}{\emph{International Conference on Learning
  Representations}}.
\newblock


\bibitem[Haeberli(1990)]%
        {10.1145/97880.97902}
\bibfield{author}{\bibinfo{person}{Paul Haeberli}.}
  \bibinfo{year}{1990}\natexlab{}.
\newblock \showarticletitle{Paint by Numbers: Abstract Image Representations}.
\newblock \bibinfo{journal}{\emph{SIGGRAPH Comput. Graph.}}
  \bibinfo{volume}{24}, \bibinfo{number}{4} (\bibinfo{date}{Sept.}
  \bibinfo{year}{1990}), \bibinfo{pages}{207–214}.
\newblock
\showISSN{0097-8930}
\urldef\tempurl%
\url{https://doi.org/10.1145/97880.97902}
\showDOI{\tempurl}


\bibitem[Hausner(2001)]%
        {10.1145/383259.383327}
\bibfield{author}{\bibinfo{person}{Alejo Hausner}.}
  \bibinfo{year}{2001}\natexlab{}.
\newblock \showarticletitle{Simulating Decorative Mosaics}. In
  \bibinfo{booktitle}{\emph{Proceedings of the 28th Annual Conference on
  Computer Graphics and Interactive Techniques}}
  \emph{(\bibinfo{series}{SIGGRAPH '01})}. \bibinfo{publisher}{Association for
  Computing Machinery}, \bibinfo{address}{New York, NY, USA},
  \bibinfo{pages}{573–580}.
\newblock
\showISBNx{158113374X}
\urldef\tempurl%
\url{https://doi.org/10.1145/383259.383327}
\showDOI{\tempurl}


\bibitem[Hays and Essa(2004)]%
        {10.1145/987657.987676}
\bibfield{author}{\bibinfo{person}{James Hays} {and} \bibinfo{person}{Irfan
  Essa}.} \bibinfo{year}{2004}\natexlab{}.
\newblock \showarticletitle{Image and Video Based Painterly Animation}. In
  \bibinfo{booktitle}{\emph{Proceedings of the 3rd International Symposium on
  Non-Photorealistic Animation and Rendering}} (Annecy, France)
  \emph{(\bibinfo{series}{NPAR '04})}. \bibinfo{publisher}{Association for
  Computing Machinery}, \bibinfo{address}{New York, NY, USA},
  \bibinfo{pages}{113–120}.
\newblock
\showISBNx{1581138873}
\urldef\tempurl%
\url{https://doi.org/10.1145/987657.987676}
\showDOI{\tempurl}


\bibitem[Healey(2001)]%
        {healey2001formalizing}
\bibfield{author}{\bibinfo{person}{Christopher~G Healey}.}
  \bibinfo{year}{2001}\natexlab{}.
\newblock \showarticletitle{Formalizing artistic techniques and scientific
  visualization for painted renditions of complex information spaces}. In
  \bibinfo{booktitle}{\emph{IJCAI}}, Vol.~\bibinfo{volume}{1}.
  \bibinfo{pages}{371--376}.
\newblock


\bibitem[Hertzmann(1998)]%
        {hertzmann1998painterly}
\bibfield{author}{\bibinfo{person}{Aaron Hertzmann}.}
  \bibinfo{year}{1998}\natexlab{}.
\newblock \showarticletitle{Painterly rendering with curved brush strokes of
  multiple sizes}. In \bibinfo{booktitle}{\emph{Proceedings of the 25th annual
  conference on Computer graphics and interactive techniques}}.
  \bibinfo{pages}{453--460}.
\newblock


\bibitem[Hertzmann(2001)]%
        {hertzmann2001paint}
\bibfield{author}{\bibinfo{person}{Aaron Hertzmann}.}
  \bibinfo{year}{2001}\natexlab{}.
\newblock \showarticletitle{Paint by relaxation}. In
  \bibinfo{booktitle}{\emph{Proceedings. Computer Graphics International
  2001}}. IEEE, \bibinfo{pages}{47--54}.
\newblock


\bibitem[Hertzmann(2003)]%
        {hertzmann2003survey}
\bibfield{author}{\bibinfo{person}{Aaron Hertzmann}.}
  \bibinfo{year}{2003}\natexlab{}.
\newblock \showarticletitle{A Survey of Stroke-Based Rendering}.
\newblock \bibinfo{journal}{\emph{IEEE Computer Graphics and Applications}}
  \bibinfo{volume}{23}, \bibinfo{number}{04} (\bibinfo{year}{2003}),
  \bibinfo{pages}{70--81}.
\newblock


\bibitem[Hertzmann(2009)]%
        {hertzmann2009stroke}
\bibfield{author}{\bibinfo{person}{Aaron Hertzmann}.}
  \bibinfo{year}{2009}\natexlab{}.
\newblock \showarticletitle{Stroke-based rendering}.
\newblock \bibinfo{journal}{\emph{Recent Advances in Npr for Art \&
  Visualization}}  \bibinfo{volume}{1} (\bibinfo{year}{2009}).
\newblock


\bibitem[Hiller et~al\mbox{.}(2003)]%
        {hiller2003beyond}
\bibfield{author}{\bibinfo{person}{Stefan Hiller}, \bibinfo{person}{Heino
  Hellwig}, {and} \bibinfo{person}{Oliver Deussen}.}
  \bibinfo{year}{2003}\natexlab{}.
\newblock \showarticletitle{Beyond stippling—Methods for distributing objects
  on the plane}. In \bibinfo{booktitle}{\emph{Computer Graphics Forum}},
  Vol.~\bibinfo{volume}{22}. Wiley Online Library, \bibinfo{pages}{515--522}.
\newblock
Issue 3.


\bibitem[Huang and Belongie(2017)]%
        {Huang_2017_ICCV}
\bibfield{author}{\bibinfo{person}{Xun Huang} {and} \bibinfo{person}{Serge
  Belongie}.} \bibinfo{year}{2017}\natexlab{}.
\newblock \showarticletitle{Arbitrary Style Transfer in Real-Time With Adaptive
  Instance Normalization}. In \bibinfo{booktitle}{\emph{Proceedings of the IEEE
  International Conference on Computer Vision (ICCV)}}.
\newblock


\bibitem[Huang et~al\mbox{.}(2019)]%
        {9010329}
\bibfield{author}{\bibinfo{person}{Zhewei Huang}, \bibinfo{person}{Shuchang
  Zhou}, {and} \bibinfo{person}{Wen Heng}.} \bibinfo{year}{2019}\natexlab{}.
\newblock \showarticletitle{Learning to Paint With Model-Based Deep
  Reinforcement Learning}. In \bibinfo{booktitle}{\emph{2019 IEEE/CVF
  International Conference on Computer Vision (ICCV)}}.
  \bibinfo{pages}{8708--8717}.
\newblock
\urldef\tempurl%
\url{https://doi.org/10.1109/ICCV.2019.00880}
\showDOI{\tempurl}


\bibitem[Jobard and Lefer(1997)]%
        {jobard1997creating}
\bibfield{author}{\bibinfo{person}{Bruno Jobard} {and} \bibinfo{person}{Wilfrid
  Lefer}.} \bibinfo{year}{1997}\natexlab{}.
\newblock \showarticletitle{Creating evenly-spaced streamlines of arbitrary
  density}.
\newblock In \bibinfo{booktitle}{\emph{Visualization in Scientific
  Computing’97}}. \bibinfo{publisher}{Springer}, \bibinfo{pages}{43--55}.
\newblock


\bibitem[Kang et~al\mbox{.}(2007)]%
        {10.1145/1274871.1274878}
\bibfield{author}{\bibinfo{person}{Henry Kang}, \bibinfo{person}{Seungyong
  Lee}, {and} \bibinfo{person}{Charles~K. Chui}.}
  \bibinfo{year}{2007}\natexlab{}.
\newblock \showarticletitle{Coherent Line Drawing}. In
  \bibinfo{booktitle}{\emph{Proceedings of the 5th International Symposium on
  Non-Photorealistic Animation and Rendering}} (San Diego, California)
  \emph{(\bibinfo{series}{NPAR '07})}. \bibinfo{publisher}{Association for
  Computing Machinery}, \bibinfo{address}{New York, NY, USA},
  \bibinfo{pages}{43–50}.
\newblock
\showISBNx{9781595936240}
\urldef\tempurl%
\url{https://doi.org/10.1145/1274871.1274878}
\showDOI{\tempurl}


\bibitem[Kang et~al\mbox{.}(2009)]%
        {4522547}
\bibfield{author}{\bibinfo{person}{Henry Kang}, \bibinfo{person}{Seungyong
  Lee}, {and} \bibinfo{person}{Charles~K. Chui}.}
  \bibinfo{year}{2009}\natexlab{}.
\newblock \showarticletitle{Flow-Based Image Abstraction}.
\newblock \bibinfo{journal}{\emph{IEEE Transactions on Visualization and
  Computer Graphics}} \bibinfo{volume}{15}, \bibinfo{number}{1}
  (\bibinfo{year}{2009}), \bibinfo{pages}{62--76}.
\newblock
\urldef\tempurl%
\url{https://doi.org/10.1109/TVCG.2008.81}
\showDOI{\tempurl}


\bibitem[Kang et~al\mbox{.}(2006)]%
        {kang2006unified}
\bibfield{author}{\bibinfo{person}{Hyung~W Kang}, \bibinfo{person}{Charles~K
  Chui}, {and} \bibinfo{person}{Uday~K Chakraborty}.}
  \bibinfo{year}{2006}\natexlab{}.
\newblock \showarticletitle{A unified scheme for adaptive stroke-based
  rendering}.
\newblock \bibinfo{journal}{\emph{The Visual Computer}} \bibinfo{volume}{22},
  \bibinfo{number}{9} (\bibinfo{year}{2006}), \bibinfo{pages}{814--824}.
\newblock


\bibitem[Kim et~al\mbox{.}(2008)]%
        {10.1111/j.1467-8659.2008.01259.x}
\bibfield{author}{\bibinfo{person}{Dongyeon Kim}, \bibinfo{person}{Minjung
  Son}, \bibinfo{person}{Yunjin Lee}, \bibinfo{person}{Henry Kang}, {and}
  \bibinfo{person}{Seungyong Lee}.} \bibinfo{year}{2008}\natexlab{}.
\newblock \showarticletitle{Feature-Guided Image Stippling}. In
  \bibinfo{booktitle}{\emph{Proceedings of the Nineteenth Eurographics
  Conference on Rendering}} (Sarajevo, Bosnia and Herzegovina)
  \emph{(\bibinfo{series}{EGSR '08})}. \bibinfo{publisher}{Eurographics
  Association}, \bibinfo{address}{Goslar, DEU}, \bibinfo{pages}{1209–1216}.
\newblock
\urldef\tempurl%
\url{https://doi.org/10.1111/j.1467-8659.2008.01259.x}
\showDOI{\tempurl}


\bibitem[Kim and Pellacini(2002)]%
        {10.1145/566570.566633}
\bibfield{author}{\bibinfo{person}{Junhwan Kim} {and} \bibinfo{person}{Fabio
  Pellacini}.} \bibinfo{year}{2002}\natexlab{}.
\newblock \showarticletitle{Jigsaw Image Mosaics}. In
  \bibinfo{booktitle}{\emph{Proceedings of the 29th Annual Conference on
  Computer Graphics and Interactive Techniques}} (San Antonio, Texas)
  \emph{(\bibinfo{series}{SIGGRAPH '02})}. \bibinfo{publisher}{Association for
  Computing Machinery}, \bibinfo{address}{New York, NY, USA},
  \bibinfo{pages}{657–664}.
\newblock
\showISBNx{1581135211}
\urldef\tempurl%
\url{https://doi.org/10.1145/566570.566633}
\showDOI{\tempurl}


\bibitem[Kim et~al\mbox{.}(2009)]%
        {10.1145/1572614.1572622}
\bibfield{author}{\bibinfo{person}{Sung~Ye Kim}, \bibinfo{person}{Ross
  Maciejewski}, \bibinfo{person}{Tobias Isenberg}, \bibinfo{person}{William~M.
  Andrews}, \bibinfo{person}{Wei Chen}, \bibinfo{person}{Mario~Costa Sousa},
  {and} \bibinfo{person}{David~S. Ebert}.} \bibinfo{year}{2009}\natexlab{}.
\newblock \showarticletitle{Stippling by Example}. In
  \bibinfo{booktitle}{\emph{Proceedings of the 7th International Symposium on
  Non-Photorealistic Animation and Rendering}} (New Orleans, Louisiana)
  \emph{(\bibinfo{series}{NPAR '09})}. \bibinfo{publisher}{Association for
  Computing Machinery}, \bibinfo{address}{New York, NY, USA},
  \bibinfo{pages}{41–50}.
\newblock
\showISBNx{9781605586045}
\urldef\tempurl%
\url{https://doi.org/10.1145/1572614.1572622}
\showDOI{\tempurl}


\bibitem[Kotovenko et~al\mbox{.}(2021)]%
        {Kotovenko_2021_CVPR}
\bibfield{author}{\bibinfo{person}{Dmytro Kotovenko}, \bibinfo{person}{Matthias
  Wright}, \bibinfo{person}{Arthur Heimbrecht}, {and} \bibinfo{person}{Bjorn
  Ommer}.} \bibinfo{year}{2021}\natexlab{}.
\newblock \showarticletitle{Rethinking Style Transfer: From Pixels to
  Parameterized Brushstrokes}. In \bibinfo{booktitle}{\emph{Proceedings of the
  IEEE/CVF Conference on Computer Vision and Pattern Recognition (CVPR)}}.
  \bibinfo{pages}{12196--12205}.
\newblock


\bibitem[Laidlaw(2001)]%
        {laidlaw2001loose}
\bibfield{author}{\bibinfo{person}{David~H. Laidlaw}.}
  \bibinfo{year}{2001}\natexlab{}.
\newblock \showarticletitle{Loose, artistic" textures" for visualization}.
\newblock \bibinfo{journal}{\emph{IEEE Computer Graphics and Applications}}
  \bibinfo{volume}{21}, \bibinfo{number}{2} (\bibinfo{year}{2001}),
  \bibinfo{pages}{6--9}.
\newblock


\bibitem[Lin et~al\mbox{.}(2010)]%
        {10.1145/1809939.1809948}
\bibfield{author}{\bibinfo{person}{Liang Lin}, \bibinfo{person}{Kun Zeng},
  \bibinfo{person}{Han Lv}, \bibinfo{person}{Yizhou Wang},
  \bibinfo{person}{Yingqing Xu}, {and} \bibinfo{person}{Song-Chun Zhu}.}
  \bibinfo{year}{2010}\natexlab{}.
\newblock \showarticletitle{Painterly Animation Using Video Semantics and
  Feature Correspondence}. In \bibinfo{booktitle}{\emph{Proceedings of the 8th
  International Symposium on Non-Photorealistic Animation and Rendering}}
  (Annecy, France) \emph{(\bibinfo{series}{NPAR '10})}.
  \bibinfo{publisher}{Association for Computing Machinery},
  \bibinfo{address}{New York, NY, USA}, \bibinfo{pages}{73–80}.
\newblock
\showISBNx{9781450301251}
\urldef\tempurl%
\url{https://doi.org/10.1145/1809939.1809948}
\showDOI{\tempurl}


\bibitem[Lindemeier et~al\mbox{.}(2015)]%
        {lindemeier2015hardware}
\bibfield{author}{\bibinfo{person}{Thomas Lindemeier}, \bibinfo{person}{Jens
  Metzner}, \bibinfo{person}{Lena Pollak}, {and} \bibinfo{person}{Oliver
  Deussen}.} \bibinfo{year}{2015}\natexlab{}.
\newblock \showarticletitle{Hardware-Based Non-Photorealistic Rendering Using a
  Painting Robot}. In \bibinfo{booktitle}{\emph{Computer graphics forum}},
  Vol.~\bibinfo{volume}{34}. Wiley Online Library, \bibinfo{pages}{311--323}.
\newblock


\bibitem[Litwinowicz(1997)]%
        {10.1145/258734.258893}
\bibfield{author}{\bibinfo{person}{Peter Litwinowicz}.}
  \bibinfo{year}{1997}\natexlab{}.
\newblock \showarticletitle{Processing Images and Video for an Impressionist
  Effect}. In \bibinfo{booktitle}{\emph{Proceedings of the 24th Annual
  Conference on Computer Graphics and Interactive Techniques}}
  \emph{(\bibinfo{series}{SIGGRAPH '97})}. \bibinfo{publisher}{ACM
  Press/Addison-Wesley Publishing Co.}, \bibinfo{address}{USA},
  \bibinfo{pages}{407–414}.
\newblock
\showISBNx{0897918967}
\urldef\tempurl%
\url{https://doi.org/10.1145/258734.258893}
\showDOI{\tempurl}


\bibitem[Liu et~al\mbox{.}(2021)]%
        {liu2021paint}
\bibfield{author}{\bibinfo{person}{Songhua Liu}, \bibinfo{person}{Tianwei Lin},
  \bibinfo{person}{Dongliang He}, \bibinfo{person}{Fu Li},
  \bibinfo{person}{Ruifeng Deng}, \bibinfo{person}{Xin Li},
  \bibinfo{person}{Errui Ding}, {and} \bibinfo{person}{Hao Wang}.}
  \bibinfo{year}{2021}\natexlab{}.
\newblock \showarticletitle{Paint transformer: Feed forward neural painting
  with stroke prediction}. In \bibinfo{booktitle}{\emph{Proceedings of the
  IEEE/CVF International Conference on Computer Vision}}.
  \bibinfo{pages}{6598--6607}.
\newblock


\bibitem[Liu et~al\mbox{.}(2006)]%
        {4015453}
\bibfield{author}{\bibinfo{person}{Zhanping Liu}, \bibinfo{person}{Robert
  Moorhead}, {and} \bibinfo{person}{Joe Groner}.}
  \bibinfo{year}{2006}\natexlab{}.
\newblock \showarticletitle{An Advanced Evenly-Spaced Streamline Placement
  Algorithm}.
\newblock \bibinfo{journal}{\emph{IEEE Transactions on Visualization and
  Computer Graphics}} \bibinfo{volume}{12}, \bibinfo{number}{5}
  (\bibinfo{year}{2006}), \bibinfo{pages}{965--972}.
\newblock
\urldef\tempurl%
\url{https://doi.org/10.1109/TVCG.2006.116}
\showDOI{\tempurl}


\bibitem[Lloyd(1982)]%
        {lloyd1982least}
\bibfield{author}{\bibinfo{person}{Stuart Lloyd}.}
  \bibinfo{year}{1982}\natexlab{}.
\newblock \showarticletitle{Least squares quantization in PCM}.
\newblock \bibinfo{journal}{\emph{IEEE transactions on information theory}}
  \bibinfo{volume}{28}, \bibinfo{number}{2} (\bibinfo{year}{1982}),
  \bibinfo{pages}{129--137}.
\newblock


\bibitem[Lu et~al\mbox{.}(2002)]%
        {1183777}
\bibfield{author}{\bibinfo{person}{Aidong Lu}, \bibinfo{person}{C.J. Morris},
  \bibinfo{person}{D.S. Ebert}, \bibinfo{person}{P. Rheingans}, {and}
  \bibinfo{person}{C. Hansen}.} \bibinfo{year}{2002}\natexlab{}.
\newblock \showarticletitle{Non-photorealistic volume rendering using stippling
  techniques}. In \bibinfo{booktitle}{\emph{IEEE Visualization, 2002. VIS
  2002.}} \bibinfo{pages}{211--218}.
\newblock
\urldef\tempurl%
\url{https://doi.org/10.1109/VISUAL.2002.1183777}
\showDOI{\tempurl}


\bibitem[Ma et~al\mbox{.}(2018)]%
        {ma2018incremental}
\bibfield{author}{\bibinfo{person}{Lei Ma}, \bibinfo{person}{Yanyun Chen},
  \bibinfo{person}{Yinling Qian}, {and} \bibinfo{person}{Hanqiu Sun}.}
  \bibinfo{year}{2018}\natexlab{}.
\newblock \showarticletitle{Incremental Voronoi sets for instant stippling}.
\newblock \bibinfo{journal}{\emph{The Visual Computer}} \bibinfo{volume}{34},
  \bibinfo{number}{6} (\bibinfo{year}{2018}), \bibinfo{pages}{863--873}.
\newblock


\bibitem[Mackay(1998)]%
        {mackay1998introduction}
\bibfield{author}{\bibinfo{person}{David John~Cameron Mackay}.}
  \bibinfo{year}{1998}\natexlab{}.
\newblock \showarticletitle{Introduction to monte carlo methods}.
\newblock In \bibinfo{booktitle}{\emph{Learning in graphical models}}.
  \bibinfo{publisher}{Springer}, \bibinfo{pages}{175--204}.
\newblock


\bibitem[Mattausch et~al\mbox{.}(2003)]%
        {10.1145/984952.984987}
\bibfield{author}{\bibinfo{person}{Oliver Mattausch}, \bibinfo{person}{Thomas
  Theu\ss{}l}, \bibinfo{person}{Helwig Hauser}, {and} \bibinfo{person}{Eduard
  Gr\"{o}ller}.} \bibinfo{year}{2003}\natexlab{}.
\newblock \showarticletitle{Strategies for Interactive Exploration of 3D Flow
  Using Evenly-Spaced Illuminated Streamlines}. In
  \bibinfo{booktitle}{\emph{Proceedings of the 19th Spring Conference on
  Computer Graphics}} (Budmerice, Slovakia) \emph{(\bibinfo{series}{SCCG
  '03})}. \bibinfo{publisher}{Association for Computing Machinery},
  \bibinfo{address}{New York, NY, USA}, \bibinfo{pages}{213–222}.
\newblock
\showISBNx{158113861X}
\urldef\tempurl%
\url{https://doi.org/10.1145/984952.984987}
\showDOI{\tempurl}


\bibitem[Pastor et~al\mbox{.}(2003)]%
        {1210866}
\bibfield{author}{\bibinfo{person}{O.M. Pastor}, \bibinfo{person}{B.
  Freudenberg}, {and} \bibinfo{person}{T. Strothotte}.}
  \bibinfo{year}{2003}\natexlab{}.
\newblock \showarticletitle{Real-time animated stippling}.
\newblock \bibinfo{journal}{\emph{IEEE Computer Graphics and Applications}}
  \bibinfo{volume}{23}, \bibinfo{number}{4} (\bibinfo{year}{2003}),
  \bibinfo{pages}{62--68}.
\newblock
\urldef\tempurl%
\url{https://doi.org/10.1109/MCG.2003.1210866}
\showDOI{\tempurl}


\bibitem[Salisbury et~al\mbox{.}(1996)]%
        {salisbury1996scale}
\bibfield{author}{\bibinfo{person}{Mike Salisbury}, \bibinfo{person}{Corin
  Anderson}, \bibinfo{person}{Dani Lischinski}, {and} \bibinfo{person}{David~H
  Salesin}.} \bibinfo{year}{1996}\natexlab{}.
\newblock \showarticletitle{Scale-dependent reproduction of pen-and-ink
  illustrations}. In \bibinfo{booktitle}{\emph{Proceedings of the 23rd annual
  conference on Computer graphics and interactive techniques}}.
  \bibinfo{pages}{461--468}.
\newblock


\bibitem[Salisbury et~al\mbox{.}(1994)]%
        {10.1145/192161.192185}
\bibfield{author}{\bibinfo{person}{Michael~P. Salisbury},
  \bibinfo{person}{Sean~E. Anderson}, \bibinfo{person}{Ronen Barzel}, {and}
  \bibinfo{person}{David~H. Salesin}.} \bibinfo{year}{1994}\natexlab{}.
\newblock \showarticletitle{Interactive Pen-and-Ink Illustration}. In
  \bibinfo{booktitle}{\emph{Proceedings of the 21st Annual Conference on
  Computer Graphics and Interactive Techniques}}
  \emph{(\bibinfo{series}{SIGGRAPH '94})}. \bibinfo{publisher}{Association for
  Computing Machinery}, \bibinfo{address}{New York, NY, USA},
  \bibinfo{pages}{101–108}.
\newblock
\showISBNx{0897916670}
\urldef\tempurl%
\url{https://doi.org/10.1145/192161.192185}
\showDOI{\tempurl}


\bibitem[Salisbury et~al\mbox{.}(1997)]%
        {10.1145/258734.258890}
\bibfield{author}{\bibinfo{person}{Michael~P. Salisbury},
  \bibinfo{person}{Michael~T. Wong}, \bibinfo{person}{John~F. Hughes}, {and}
  \bibinfo{person}{David~H. Salesin}.} \bibinfo{year}{1997}\natexlab{}.
\newblock \showarticletitle{Orientable Textures for Image-Based Pen-and-Ink
  Illustration}. In \bibinfo{booktitle}{\emph{Proceedings of the 24th Annual
  Conference on Computer Graphics and Interactive Techniques}}
  \emph{(\bibinfo{series}{SIGGRAPH '97})}. \bibinfo{publisher}{ACM
  Press/Addison-Wesley Publishing Co.}, \bibinfo{address}{USA},
  \bibinfo{pages}{401–406}.
\newblock
\showISBNx{0897918967}
\urldef\tempurl%
\url{https://doi.org/10.1145/258734.258890}
\showDOI{\tempurl}


\bibitem[Secord(2002)]%
        {10.1145/508530.508537}
\bibfield{author}{\bibinfo{person}{Adrian Secord}.}
  \bibinfo{year}{2002}\natexlab{}.
\newblock \showarticletitle{Weighted Voronoi Stippling}. In
  \bibinfo{booktitle}{\emph{Proceedings of the 2nd International Symposium on
  Non-Photorealistic Animation and Rendering}} (Annecy, France)
  \emph{(\bibinfo{series}{NPAR '02})}. \bibinfo{publisher}{Association for
  Computing Machinery}, \bibinfo{address}{New York, NY, USA},
  \bibinfo{pages}{37–43}.
\newblock
\showISBNx{1581134940}
\urldef\tempurl%
\url{https://doi.org/10.1145/508530.508537}
\showDOI{\tempurl}


\bibitem[Silvers(1996)]%
        {silvers1996photomosaics}
\bibfield{author}{\bibinfo{person}{Robert Silvers}.}
  \bibinfo{year}{1996}\natexlab{}.
\newblock \emph{\bibinfo{title}{Photomosaics: putting pictures in their
  place}}.
\newblock \bibinfo{thesistype}{Ph.\,D. Dissertation}.
  \bibinfo{school}{Massachusetts Institute of Technology}.
\newblock


\bibitem[Tong et~al\mbox{.}(2021)]%
        {Tong_Chen_Ni_Wang_2021}
\bibfield{author}{\bibinfo{person}{Zhengyan Tong}, \bibinfo{person}{Xuanhong
  Chen}, \bibinfo{person}{Bingbing Ni}, {and} \bibinfo{person}{Xiaohang Wang}.}
  \bibinfo{year}{2021}\natexlab{}.
\newblock \showarticletitle{Sketch Generation with Drawing Process Guided by
  Vector Flow and Grayscale}.
\newblock \bibinfo{journal}{\emph{Proceedings of the AAAI Conference on
  Artificial Intelligence}} \bibinfo{volume}{35}, \bibinfo{number}{1}
  (\bibinfo{date}{May} \bibinfo{year}{2021}), \bibinfo{pages}{609--616}.
\newblock
\urldef\tempurl%
\url{https://ojs.aaai.org/index.php/AAAI/article/view/16140}
\showURL{%
\tempurl}


\bibitem[Turk and Banks(1996)]%
        {turk1996image}
\bibfield{author}{\bibinfo{person}{Greg Turk} {and} \bibinfo{person}{David
  Banks}.} \bibinfo{year}{1996}\natexlab{}.
\newblock \showarticletitle{Image-guided streamline placement}. In
  \bibinfo{booktitle}{\emph{Proceedings of the 23rd annual conference on
  Computer graphics and interactive techniques}}. \bibinfo{pages}{453--460}.
\newblock


\bibitem[Van~Laerhoven et~al\mbox{.}(2004)]%
        {1309281}
\bibfield{author}{\bibinfo{person}{T. Van~Laerhoven}, \bibinfo{person}{J.
  Liesenborgs}, {and} \bibinfo{person}{F. Van~Reeth}.}
  \bibinfo{year}{2004}\natexlab{}.
\newblock \showarticletitle{Real-time watercolor painting on a distributed
  paper model}. In \bibinfo{booktitle}{\emph{Proceedings Computer Graphics
  International, 2004.}} \bibinfo{pages}{640--643}.
\newblock
\urldef\tempurl%
\url{https://doi.org/10.1109/CGI.2004.1309281}
\showDOI{\tempurl}


\bibitem[Vanderhaeghe et~al\mbox{.}(2007)]%
        {vanderhaeghe2007dynamic}
\bibfield{author}{\bibinfo{person}{David Vanderhaeghe}, \bibinfo{person}{Pascal
  Barla}, \bibinfo{person}{Jo{\"e}lle Thollot}, {and}
  \bibinfo{person}{Fran{\c{c}}ois~X Sillion}.} \bibinfo{year}{2007}\natexlab{}.
\newblock \showarticletitle{Dynamic point distribution for stroke-based
  rendering}. In \bibinfo{booktitle}{\emph{Eurogaphics Symposium on
  Rendering}}. Eurographics Association, \bibinfo{pages}{139--146}.
\newblock


\bibitem[Winkenbach and Salesin(1994)]%
        {10.1145/192161.192184}
\bibfield{author}{\bibinfo{person}{Georges Winkenbach} {and}
  \bibinfo{person}{David~H. Salesin}.} \bibinfo{year}{1994}\natexlab{}.
\newblock \showarticletitle{Computer-Generated Pen-and-Ink Illustration}. In
  \bibinfo{booktitle}{\emph{Proceedings of the 21st Annual Conference on
  Computer Graphics and Interactive Techniques}}
  \emph{(\bibinfo{series}{SIGGRAPH '94})}. \bibinfo{publisher}{Association for
  Computing Machinery}, \bibinfo{address}{New York, NY, USA},
  \bibinfo{pages}{91–100}.
\newblock
\showISBNx{0897916670}
\urldef\tempurl%
\url{https://doi.org/10.1145/192161.192184}
\showDOI{\tempurl}


\bibitem[Winkenbach and Salesin(1996)]%
        {winkenbach1996rendering}
\bibfield{author}{\bibinfo{person}{Georges Winkenbach} {and}
  \bibinfo{person}{David~H Salesin}.} \bibinfo{year}{1996}\natexlab{}.
\newblock \showarticletitle{Rendering parametric surfaces in pen and ink}. In
  \bibinfo{booktitle}{\emph{Proceedings of the 23rd annual conference on
  Computer graphics and interactive techniques}}. \bibinfo{pages}{469--476}.
\newblock


\bibitem[Xie et~al\mbox{.}(2013)]%
        {xie2013artist}
\bibfield{author}{\bibinfo{person}{Ning Xie}, \bibinfo{person}{Hirotaka
  Hachiya}, {and} \bibinfo{person}{Masashi Sugiyama}.}
  \bibinfo{year}{2013}\natexlab{}.
\newblock \showarticletitle{Artist agent: A reinforcement learning approach to
  automatic stroke generation in oriental ink painting}.
\newblock \bibinfo{journal}{\emph{IEICE TRANSACTIONS on Information and
  Systems}} \bibinfo{volume}{96}, \bibinfo{number}{5} (\bibinfo{year}{2013}),
  \bibinfo{pages}{1134--1144}.
\newblock


\bibitem[Xie et~al\mbox{.}(2015)]%
        {xie2015stroke}
\bibfield{author}{\bibinfo{person}{Ning Xie}, \bibinfo{person}{Tingting Zhao},
  \bibinfo{person}{Feng Tian}, \bibinfo{person}{Xiao~Hua Zhang}, {and}
  \bibinfo{person}{Masashi Sugiyama}.} \bibinfo{year}{2015}\natexlab{}.
\newblock \showarticletitle{Stroke-based stylization learning and rendering
  with inverse reinforcement learning}. In
  \bibinfo{booktitle}{\emph{Twenty-fourth international joint conference on
  artificial intelligence}}.
\newblock


\bibitem[Zeng et~al\mbox{.}(2009)]%
        {zeng2009image}
\bibfield{author}{\bibinfo{person}{Kun Zeng}, \bibinfo{person}{Mingtian Zhao},
  \bibinfo{person}{Caiming Xiong}, {and} \bibinfo{person}{Song~Chun Zhu}.}
  \bibinfo{year}{2009}\natexlab{}.
\newblock \showarticletitle{From image parsing to painterly rendering.}
\newblock \bibinfo{journal}{\emph{ACM Trans. Graph.}} \bibinfo{volume}{29},
  \bibinfo{number}{1} (\bibinfo{year}{2009}), \bibinfo{pages}{2--1}.
\newblock


\bibitem[Zheng et~al\mbox{.}(2019)]%
        {zheng2018strokenet}
\bibfield{author}{\bibinfo{person}{Ningyuan Zheng}, \bibinfo{person}{Yifan
  Jiang}, {and} \bibinfo{person}{Dingjiang Huang}.}
  \bibinfo{year}{2019}\natexlab{}.
\newblock \showarticletitle{StrokeNet: A Neural Painting Environment}. In
  \bibinfo{booktitle}{\emph{International Conference on Learning
  Representations}}.
\newblock


\bibitem[Zhu et~al\mbox{.}(2017)]%
        {zhu2017unpaired}
\bibfield{author}{\bibinfo{person}{Jun-Yan Zhu}, \bibinfo{person}{Taesung
  Park}, \bibinfo{person}{Phillip Isola}, {and} \bibinfo{person}{Alexei~A
  Efros}.} \bibinfo{year}{2017}\natexlab{}.
\newblock \showarticletitle{Unpaired image-to-image translation using
  cycle-consistent adversarial networks}. In
  \bibinfo{booktitle}{\emph{Proceedings of the IEEE international conference on
  computer vision}}. \bibinfo{pages}{2223--2232}.
\newblock


\bibitem[Zou et~al\mbox{.}(2021)]%
        {Zou_2021_CVPR}
\bibfield{author}{\bibinfo{person}{Zhengxia Zou}, \bibinfo{person}{Tianyang
  Shi}, \bibinfo{person}{Shuang Qiu}, \bibinfo{person}{Yi Yuan}, {and}
  \bibinfo{person}{Zhenwei Shi}.} \bibinfo{year}{2021}\natexlab{}.
\newblock \showarticletitle{Stylized Neural Painting}. In
  \bibinfo{booktitle}{\emph{Proceedings of the IEEE/CVF Conference on Computer
  Vision and Pattern Recognition (CVPR)}}. \bibinfo{pages}{15689--15698}.
\newblock


\end{thebibliography}


\clearpage

\appendix


\begin{figure*}[ht]
  \centering
  \includegraphics[width=\linewidth]{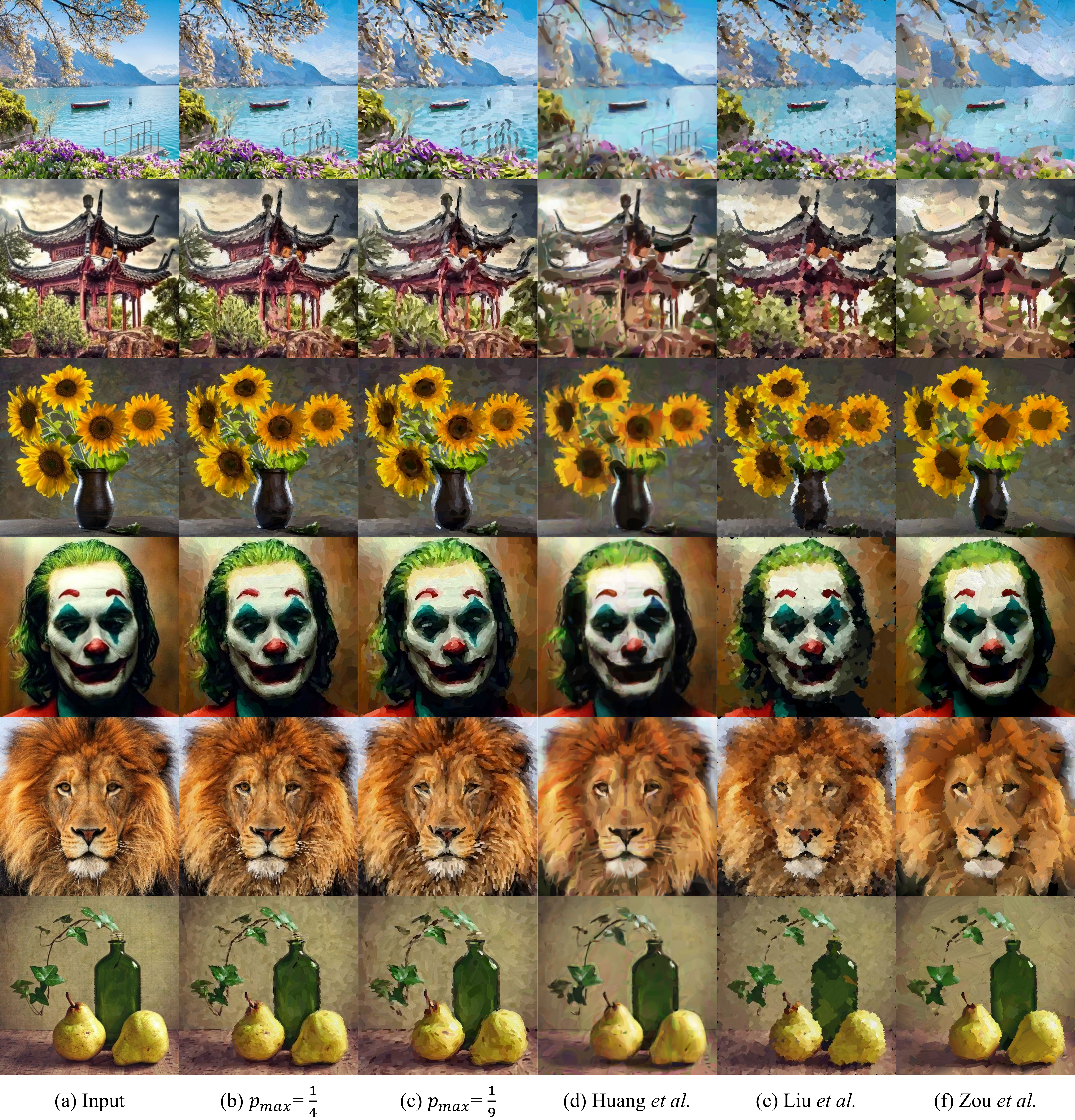}
  \caption{Oil painting comparison with three state-of-the-art methods: Huang \emph{et al.} \cite{9010329}, Liu \emph{et al.} \cite{liu2021paint}, and Zou \emph{et al.} \cite{Zou_2021_CVPR}. Column (a) is the input. Column (b) and (c) are our results using $p_{max}=\{\frac{1}{4}, \frac{1}{9}\}$, respectively. Column (d) is the result of Huang \emph{et al.} \cite{9010329}; (e) is the result of Liu \emph{et al.} \cite{liu2021paint}; (f) is the result of Zou \emph{et al.} \cite{Zou_2021_CVPR}. Please zoom in to compare details.}
  \label{fig:compare-6}
\end{figure*}

\begin{figure*}[ht]
  \centering
  \includegraphics[width=\linewidth]{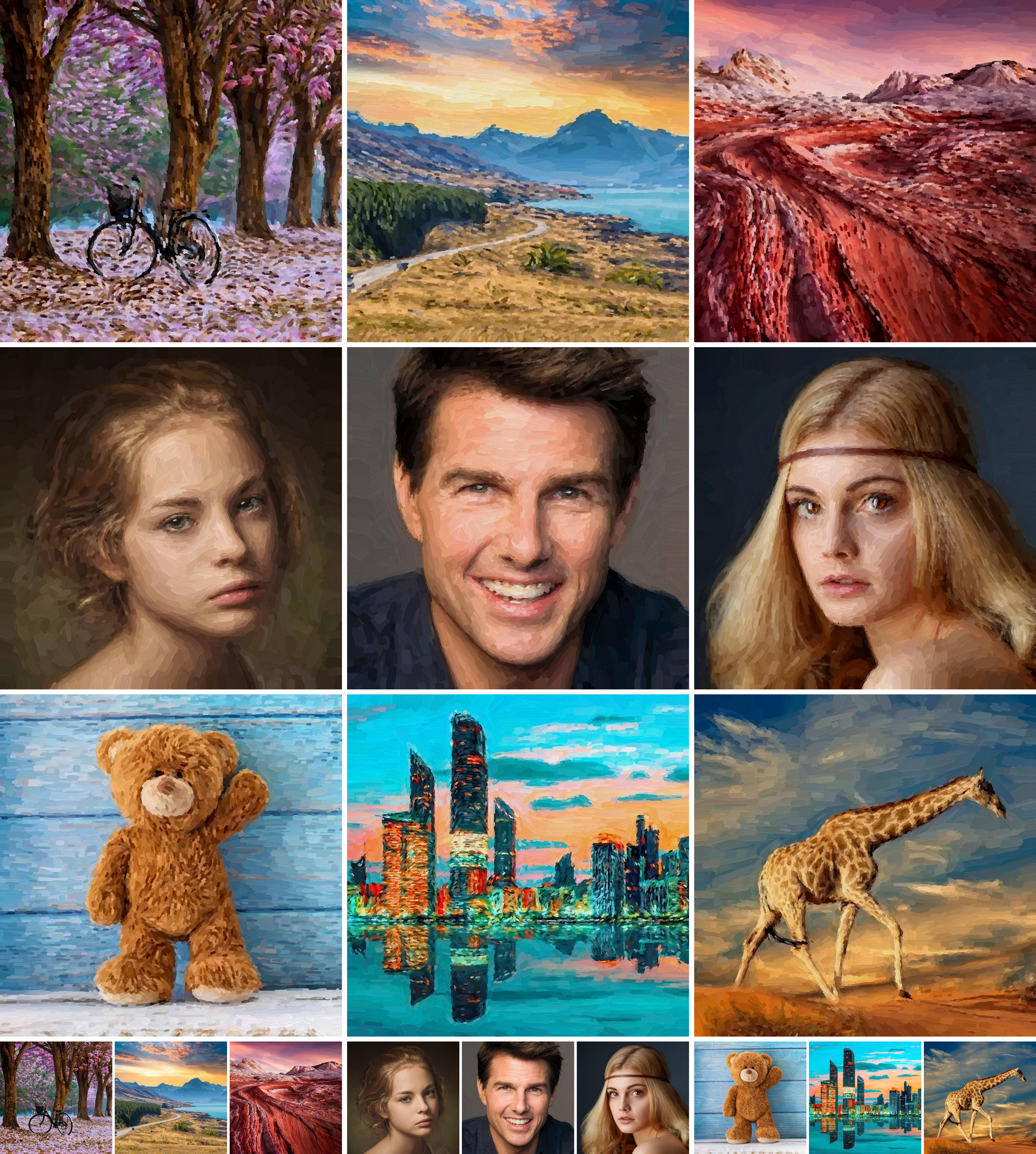}
  \caption{Nine oil paintings produced by our method using $p_{max}=\frac{1}{4}$. The input images are at the bottom.}
  \label{fig:hr}
\end{figure*}

\end{document}